\RequirePackage{amsmath}
\documentclass[runningheads,a4]{llncs}
\usepackage{graphicx}
\usepackage{amsmath,amssymb} 
\usepackage{color}
\usepackage{booktabs}
\usepackage{svg}
\usepackage{xspace}
\usepackage{hyperref}

\usepackage{tabularx}
\newcolumntype{Y}{>{\centering\arraybackslash}X}

\newcommand\subsetsim{\mathrel{%
  \ooalign{\raise0.2ex\hbox{$\subset$}\cr\hidewidth\raise-0.8ex\hbox{\scalebox{0.9}{$\sim$}}}}}
\newcommand{\etal}{\emph{et al.}\xspace}
\newcommand{\ie}{\emph{i.e.}}
\newcommand{\eg}{\emph{e.g.}}
\newcommand{\psec}[1]{\autoref{#1}}
\usepackage{longtable}

\newcommand{\netname}{S2SNet\xspace}  
\newcommand{\netnames}{S2SNets\xspace}  
\newcommand{\autoe}[1]{\mathbb{A}_{#1}}
\newcommand{\clf}[1]{#1}
\newcommand{\nn}[2]{\clf{#2}\circ\autoe{#1}}
\newcommand{\pert}[2]{\mathcal{P}_{#1,#2}}
\newcommand{\wb}{White-Box\xspace}
\newcommand{\wba}{White-Box$^+$\xspace}
\newcommand{\gm}{Gray-Box$^-$\xspace}
\newcommand{\ltwo}{$\mathcal{L}_2$\xspace}
\newcommand{\fog}{f\circ g}

\usepackage[
  detect-weight,
  exponent-product=\cdot,
  locale = DE,
  group-separator=.,
]{siunitx}
\DeclareSIUnit{\pp}{\textup{p.p.}}

\usepackage{marvosym}
\def\imp{\rightarrow}
\newenvironment{level}%
{\addtolength{\itemindent}{2em}}%
{\addtolength{\itemindent}{-2em}}

\mainmatter
\def\ECCV18SubNumber{0000}  

\title{Adversarial Defense based on Structure-to-Signal Autoencoders} 

\titlerunning{} 

\authorrunning{Palacio S., Folz J., \etal} 

\author{Joachim Folz\thanks{Authors contributed equally} \and Sebastian Palacio$^\star$ \and Joern Hees \and Damian Borth \and Andreas Dengel}
\institute{German Research Center for Artificial Intelligence (DFKI) \and TU Kaiserslautern\\ \email{first.last@dfki.de}}

\begin{document}
\maketitle

\begin{abstract}
Adversarial attack methods have demonstrated the fragility of deep neural networks.
Their imperceptible perturbations are frequently able fool classifiers into potentially dangerous misclassifications.
We propose a novel way to interpret adversarial perturbations in terms of the effective input signal that classifiers actually use.
Based on this, we apply specially trained autoencoders, referred to as \netnames, as defense mechanism.
They follow a two-stage training scheme: first unsupervised, followed by a fine-tuning of the decoder, using gradients from an existing classifier.
\netnames induce a shift in the distribution of gradients propagated through them, stripping them from class-dependent signal.
We analyze their robustness against several white-box and gray-box scenarios on the large ImageNet dataset.
Our approach reaches comparable resilience in white-box attack scenarios as other state-of-the-art defenses in gray-box scenarios.
We further analyze the relationships of AlexNet, VGG 16, ResNet 50 and Inception v3 in adversarial space, and found that VGG 16 is the easiest to fool, while perturbations from ResNet 50 are the most transferable.

\keywords{adversarial defense, CNNs, deep learning, autoencoders}
\end{abstract}

\section{Introduction}

We nowadays see an increasing adoption of deep learning techniques in production systems, partially even safety-relevant ones (as exemplified in \cite{papernot2016practical}).
Hence, it is not surprising that the discovery of adversarial spaces for neural networks~\cite{szegedy2013intriguing} has sparked a lot of interest.
A growing community has quickly focused on different ways to reach those spaces~\cite{goodfellow2014explaining,moosavi2016deepfool,carlini2017towards}, understand their properties~\cite{papernot2016practical,kurakin2016adversarial,moosavi2017universal}, and protect vulnerable models from their malicious nature~\cite{papernot2016distillation,feinman2017detecting,madry2018towards}.

Most prevalent methods exploiting adversarial spaces~\cite{goodfellow2014explaining,moosavi2016deepfool,carlini2017towards,madry2018towards} use gradients as their main starting point to search for perturbations surrounding a clean sample.
Due to the intractability of transformations modeled by neural networks and the limited amount of change that is allowed for perturbations to be considered adversarial, gradients from a classifier expose just enough information about which parts of the input correlate highly with the label that a model associates to it.
With this in mind, defenses against adversarial attacks have been devised upon those same gradients in two fundamental ways:
1) They complement traditional optimization schemes with a two-fold objective that minimizes the overall prediction cost while maximizing the perturbation space around clean images that classifiers can withstand~\cite{madry2018towards,sinha2018certifiable,tramer2018ensemble}.
2) Gradients are blocked or obfuscated in such a way that attacking algorithms can no longer use them to find effective adversarial perturbations~\cite{buckman2018thermometer,guo2018countering,song2018pixeldefend}.
Type 1 methods enjoy mathematical rigor and hence, provide formal guarantees with respect to the kind of perturbations they are robust to.
However, note that this can also be disadvantageous since networks become attack-dependent.
Any other strategy for finding perturbations could circumvent such a defense mechanism~\cite{athalye2018obfuscated}.
While effective for small-scale problems such as MNIST and CIFAR, we found no empirical evidence that these methods scale to larger problems such as ImageNet.
It has even been shown that defenses for adversarial attacks tested on small datasets do not scale well when applied to bigger problems~\cite{kurakin2016adversarial}.

Currently, large scale state-of-the-art defenses rely on the second use of gradients: suppression~\cite{liao2017defense} and blockage~\cite{guo2018countering,xie2018mitigating}.
As defined by Athalye \etal~\cite{athalye2018obfuscated}, gradient obfuscation is the result of instabilities from vanishing or exploding gradients, and the use of stochastic or non-differentiable preprocessing steps.
All these alternatives can be modeled as lossy identities where the original signal contained in the input is preserved, while the adversarial perturbation is destroyed.
The usefulness of this principle fits well with findings from a recent study showing that image classifiers only use a small fraction of the entire signal within original input.
Therefore, a portion of its information can indeed be dropped without affecting performance~\cite{palaciofolz2018normnet}.

In this paper, we propose an alternative defense that affects the information contained in gradients by reforming its class-related signal into a structural one.
Intuitively, we learn an identity function that encodes structure and decodes only the structural parts of the input necessary for classification, dropping everything else.
To this end, an autoencoder (AE) is trained to approximate the identity function that preserves only the part of the signal that is \emph{useful} for a target classifier.
The structural information is preserved by training both encoder and decoder unsupervised, and fine-tuning only the decoder with gradients coming from an existing classifier.
By using a function that only looks at structure, gradients are devoid of any class-related information, therefore invalidating the fundamental assumptions about gradients that attackers rely on.
We call this defense a \emph{Structure-To-Signal Network} (\netname).

\subsubsection{Formal Definitions}
\begin{figure}[t]
  \scriptsize
  \includesvg[svgpath=figures/,width=\linewidth]{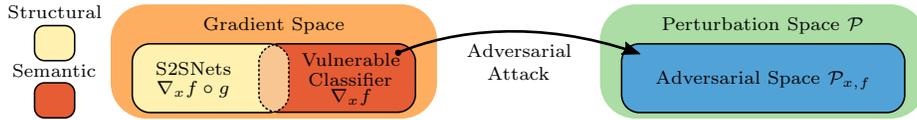}
  \caption{Overview of gradient and perturbation spaces, and their relation to \netnames and adversarial attacks.}
  \label{fig:s2sintuition}
\end{figure}

Let $ f: \mathbb{R}^3 \rightarrow \{1,\dots, k\}$ be an image classifier, and $ \tilde{x}_f $ the portion of the signal in the original input $ x $ that is effectively being used by $ f $.
In other words, $ f(x) = f(\tilde{x}_f) $ subject to $ I(x, \tilde{x}_f) < I(x, x) $, where $ I $ is a measure of information \eg, the \emph{normalized mutual information}~\cite{strehl2002cluster}.
Let $ \mathcal{P} = \{\delta: ||\delta||_L < \epsilon\}  \subseteq \mathbb{R}^3 $ be the space of all adversarial and non-adversarial perturbations under a given norm $ L \in \{1, 2, \infty\} $.
We also define the sub-space $ \pert{x}{f} \subseteq \mathcal{P} $ as the set of adversarial perturbations that are reachable from the input gradients $ \nabla_x C(f(x), y_{1\dots k}) \subseteq \mathbb{R}^3$, where $ C $ represents the cost function, and $ y_i $ is the ground truth used for training.
Note that said spaces depend on $ f $ and $ x $, but not on the class $ y $ or the cost function $ C $.
Hence, unless noted otherwise, we use the simplified notation $ \nabla_x f = \nabla_x C(f(x), y_{1\dots k})$ from now on.
Adversarial perturbations are elements $ \delta_{adv} \in \pert{x}{f} $ such that $ f(x) \neq f(x+\delta_{adv}) $ \footnote{Strictly speaking, adversarial perturbations can be reached through other domains that do not depend on gradients (as shown by Ngueyen \etal~\cite{nguyen2016synthesizing}) but so far, all instances of strong adversarial attack methods, base their entire strategy on the information of gradients.}.
As attacks use gradients to compute adversarial perturbations, and such gradients can only correspond to parts of the signal that are used by the model, it follows that $ \pert{x}{f} \equiv \pert{\tilde{x}}{f} $.
An \netname can hence be defined as a function $ g: \mathbb{R}^3 \rightarrow \mathbb{R}^3 $ such that $ g(x) = \tilde{x}_f $.
This way, $ g $ defends $ f $ by serving as a proxy for incoming, potentially malicious inputs, as well as exposing gradients via the function composition $ f \circ g $.
The relation between $ x $ and $ \tilde{x} $ concerning information implies that $ g $ produces lossy reconstructions of the input space.
We train $ g $ in a way that gradients $ \nabla_x f \circ g $ lie in a different space than those leading attackers to $ \pert{\tilde{x}}{f} $ (\autoref{fig:s2sintuition}).
In other words, we train $ g $ such that $ | \nabla_x f\circ g \cap \nabla_x f | $ is minimized.
This in turn, will cause the intersection $|\pert{x}{f \circ g} \cap \pert{\tilde{x}}{f}|$ to be smaller, resulting in perturbations that are non-adversarial (\ie, $ \delta \in \mathcal{P} - \pert{\hat{x}}{f} $).
Further details of the architecture and training of $ f $ and $ g $ are discussed in \autoref{sec:normnet}.

The architecture of this defense offers important advantages when dealing with adversarial attacks:
\begin{itemize}
  \item \textbf{Zero compromise for safe use-cases}: there is no drop in performance when the defense is deployed, but the network is not under attack.
  As the transformation done by \netname is preserving all required signal $ \tilde{x}_f $, using either $ g(x) $ or $ x $ has no impact to the classifier when clean images are used.

  \item \textbf{Removing \netname \emph{is} a defense strategy}: a novel gray-box defense (\ie, when the attacker knows the classifier but not the defense mechanism) works by giving away gradient information from an \netname to the attacker and then removing it, applying adversarial attacks (based on gradients from \netname) to the original classifier instead.
  
  \item \textbf{Attack agnostic}: \netnames rely on the same information used by adversarial attacks but not on the attacks themselves.
  Therefore, \netnames do not require any assumptions with respect to the specific way an attack works.
    
  \item \textbf{\textit{Post-hoc} implementation}: our defense uses gradients from a trained network, and can be used to defend models that are already in production.
  Likewise, no special considerations need to be made when training a new classifier from scratch.
    
  \item \textbf{Compatibility with other defenses}: due to the compositional nature of this approach, any additional defense strategies that work with the original classifier can be implemented for the ensemble $f \circ g$.

\end{itemize}

We test \netnames on two high-performing image classifiers (ResNet50~\cite{he2016deep} and Inception-v3~\cite{DBLP:journals/corr/SzegedyVISW15}) against three attack methods (Fast Gradient-Sign Method (FGSM)~\cite{goodfellow2014explaining}, Basic Iterative Method (BIM)~\cite{kurakin2016physical} and Carlini-Wager (CW)~\cite{carlini2017towards}) on the large scale ImageNet~\cite{ILSVRC15} dataset.
Experiments are conducted on both classifiers under white-box and gray-box conditions.
An evaluation of the effectiveness of \netnames with respect to regular AEs (\eg, as proposed in~\cite{meng2017magnet}) is presented in \autoref{sec:normnet}, empirically proving that \netnames are a better approximation of the signal $ \tilde{x}_f $ that is used by a classifier.
Furthermore, an evaluation of the gradient space $ \pert{x}{f} $ and $ \pert{x}{f\circ g} $ is conducted, showing that their intersection does indeed approach the empty set.

The main contributions of this paper are threefold:
First, we propose a novel way to interpret adversarial perturbations, namely in terms of the effective input signal that classifiers use $ \tilde{x}_f $.
Second, we introduce a robust and flexible defense against large-scale adversarial attacks based on \netnames.
And third, we provide a comprehensive baseline evaluation of adversarial attacks for several state-of-the-art models on a large dataset.

\section{Related Work}
\label{sec:relatedwork}

The fast growing interest in the phenomenon of adversarial attacks has gained momentum since its discovery~\cite{szegedy2013intriguing} and has had three main areas of focus.
The first area is the one that seeks new and more effective ways of reaching adversarial spaces.
In~\cite{goodfellow2014explaining}, a first comprehensive analysis of the extent of adversarial spaces was explored, proposing a fast method to compute perturbations based on the sign of gradients.
An iterative version of this method was later introduced~\cite{kurakin2016physical} and shown to work significantly better, even when applied to images that were physically printed and digitized again.
A prominent exception to attacks based on gradients succeeded using evolutionary algorithms~\cite{nguyen2016synthesizing}.
Nevertheless, this has not been a practical wide spread method, mostly due to how costly it is to compute.
Papernot \etal~\cite{papernot2016practical} showed how effective adversarial attacks could get, even with very few assumptions about the attacked model.
Finally, there are methods that go beyond a greedy iteration over the gradient space and perform different kinds of optimization that maximize misclassification while minimizing the norm of the perturbation~\cite{moosavi2016deepfool,carlini2017towards}.

The second area focuses on understanding the properties of adversarial perturbations.
The work of Goodfellow \etal~\cite{goodfellow2014explaining} was already pointing at the linear nature of neural networks as the main enabler of adversarial attacks.
This went in opposition of what was initially theorized, where the claim was that non-linearities were the main vulnerability.
Not only was it possible to perturb natural looking images to look like something completely different, but it was also possible to get models issuing predictions with high confidence using either noise images, or highly artificial patterns~\cite{szegedy2013intriguing,nguyen2016synthesizing}.
The transferability of adversarial perturbations was shown to be possible by crafting attacks on one network and using them to fool a second classifier~\cite{liu2016delving}.
However, transferable attacks are limited to the simpler methods as iterative ones tend to exploit particularities of each model, and hence lose power when used in different architectures~\cite{kurakin2016adversarial}.
As it turns out, not only is adversarial noise transferable between models but it is also possible to transfer a single universal adversarial perturbation to all samples in a dataset to achieve high misclassification rates~\cite{moosavi2017universal}.
Said individual perturbations can even be applied to physical objects and bias a model towards a specific class~\cite{brown2017adversarial}.

The third and arguably the most popular area of research has focused on how networks can be protected against such attacks.
Strategies include changing the optimization objective to account for possible adversarial spaces~\cite{madry2018towards,sinha2018certifiable}, detection~\cite{feinman2017detecting}, dataset augmentation that includes adversarial examples~\cite{goodfellow2014explaining,tramer2018ensemble}, suppressing perturbations~\cite{papernot2016distillation,song2018pixeldefend,liao2017defense,meng2017magnet,dhillon2018stochastic} or obfuscating the gradients to prevent attackers from estimating an effective perturbation~\cite{buckman2018thermometer,guo2018countering,xie2018mitigating,wang2016learning,samangouei2018defensegan}.

In this work, we build on the idea of using AEs as a compressed representation of the input~\cite{song2018pixeldefend,meng2017magnet}, but tailored towards a specific characteristic of adversarial perturbations~\cite{liao2017defense}, using the notion of useful input signal (\ie, the one effectively used by a classifier)~\cite{palaciofolz2018normnet}.
Furthermore, we explore the nature of adversarial perturbations and its relationship with network capacity, in terms of used signal.

\section{Methods}
This section explains in detail the architecture of an \netname and its particular signal-preserving training scheme, followed by an empirical evaluation of the gradients it provides.
We start by testing the robustness of \netnames in a white-box setting, and compare it to a simple baseline using regular AEs.
To recreate realistic attack conditions, we further test the ensemble network, simulating a re-parametrization technique similar to~\cite{athalye2018obfuscated}, aimed at circumventing the defense, and explore further strategies to cope with this attack.
Next, we examine the performance of \netnames in a gray-box scenario.
Finally, we provide an evaluation of the transferability of single step attacks for different models, and correlate their overlap in terms of input signal from the perspective of adversarial perturbations.

\subsection{Structure-to-Signal Networks}
\label{sec:normnet}

\netnames start out as plain AEs that are trained on the large-scale YFCC100m data set~\cite{YFCC100m}.
Only a single pass is required, as the $\approx100$ million images are more than sufficient to train the underlying SegNet architecture~\cite{badrinarayanan2015segnet2} to convergence\footnote{We choose a large architecture as an upper bound to the ideal identity function $ g(x) = \tilde{x}_f $, because it was proven capable of encoding the semantics of the deep image classifiers tested in this work.}.
This network, referred to as $\autoe{S}$, is able to reproduce input signals required by a diverse set of classifiers such that their top-1 accuracy is within $\approx 3$ percentage points of the original classification performance~\cite{palaciofolz2018normnet}.

To model the effective input signal used by a trained classifier $f$, Palacio \etal propose to further fine-tune the decoder of $\autoe{S}$ using gradients from $ f $ itself.
This allows $ \autoe{S} $ to learn a way to decode the input that retains the signal required by $f$.
This fine-tuned variant, called $\autoe{f}$, reconstructs images such that the original top-1 accuracy of $ f $ is preserved,
while amount of information in the reconstructed image (measured as normalized mutual information) decreases with respect to the original sample.

Note that, since the encoder of $ \autoe{f} $ was trained unsupervised, any intermediate representations produced by $ \autoe{f} $ are entirely class-agnostic.
This means that, during backpropagation through an \netname, gradients that can be read at the shallowest layer correspond only to information about structure.
Intuitively, gradients from $ \autoe{f} $ point to parts of the image that can be changed to influence the \emph{reconstruction} error.

In the following, we measure the extent to which gradients shift when images are forwarded through these networks.
Furthermore we explore and quantify their emerging resilience to adversarial attacks in \autoref{sec:experiments}.

\subsection{Properties of Gradient Distributions}
\label{sec:gradientnature}

To verify that the Structure-to-Signal training scheme produces large shifts in the distribution of gradients, we forward images through a classifier, a pre-trained SegNet, and through the fine-tuned counterpart to compare their gradients.
We use the magnitude of the gradients instead of raw values to stress the differences of their spatial distribution.
A large change in the position where gradients originally occur within the image is a good indicator that the information conveyed by gradients has changed.

\begin{table}[b]
\caption{Pairwise mean SSIM of input gradient magnitudes for ResNet 50 ($ R $) on the ImageNet validation set, with and without being passed through $\autoe{R}$ or $\autoe{S}$.
SSIM values of $ R $ w.r.t. any AE variant show the least similarity.
}
\centering
%
%
\begin{tabularx}{0.8\textwidth}{rYYYYYY}
\toprule 
 & & $R$ & $\nn{R}{R}$ & $\nn{S}{R}$ & $\autoe{R}$ & $\autoe{S}$ \\
\midrule
$R$         &  &  $1.00$ & $0.17$ & $0.18$ & $0.12$ & $0.14$ \\
$\nn{R}{R}$ &  &  $0.17$ & $1.00$ & $0.40$ & $0.46$ & $0.32$ \\
$\nn{S}{R}$ &  &  $0.18$ & $0.40$ & $1.00$ & $0.37$ & $0.36$ \\
$\autoe{R}$ &  &  $0.12$ & $0.46$ & $0.37$ & $1.00$ & $0.36$ \\
$\autoe{S}$ &  &  $0.14$ & $0.32$ & $0.36$ & $0.36$ & $1.00$ \\
\bottomrule
\end{tabularx}

\label{tab:gradientssim}
\end{table}

Concretely, we quantify the structural similarity~\cite{wang2004ssim} (SSIM; a locally normalized mean square error measured in a sliding window) of gradients obtained by the same image when passed through a ResNet50 ($R$), an \netname fine-tuned to defend the same ResNet50 ($\nn{R}{R}$), a plain SegNet AE coupled with ResNet50 ($\nn{S}{R}$), the fine-tuned \netname without the classifier ($\autoe{R}$), and the plain SegNet AE alone ($\autoe{S}$).
While the first three models require gradients to be computed with respect to a class label, the last two are produced by measuring the reconstruction error.
Also, note that for $\autoe{R}$ the true reconstruction cannot be directly obtained as it is indirectly defined by the classifier it was fine-tuned on.
In this case, reconstruction gradients are computed by comparing its output to the original input.

\autoref{tab:gradientssim} reports the mean SSIM of the gradients over ImageNet's validation set for all combinations of network pairs.
Note that the dissimilarity between $ \autoe{S} $ and $ \autoe{R} $ indicates that the Structure-to-Signal training scheme has indeed changed the reconstruction process \ie, the identity function being computed based on the input.
Similarly, comparing the SSIM values of $ \nn{S}{R} $ and $ \nn{R}{R} $ reveals a difference in the information contained by their gradients.
Most importantly, the similarity between the gradients coming from classifier $ R $ and either of the ensembles $ \nn{S}{R} $ and $ \nn{R}{R} $ is considerably smaller ($0.18$ and $0.17$) than any combination involving AEs exclusively ($ > 0.32 $).
This disparity indicates how much the position of gradients change (and hence the information contained within), when passed through an \netname.

Further evidence for the class-agnostic nature of gradients propagated through AEs can be found when comparing the SSIM of gradient magnitudes when different target labels are used to compute gradients.
Let $x$ be some input image and $y^*$ its true label.
We then randomly select a different label $\hat{y} \ne y^*$ and compute
\begin{equation*}
	SSIM(||\nabla_x C(f(x), y^*)||, ||\nabla_x C(f(x), \hat{y})||)
\end{equation*}
for all $x$ in ImageNet's validation set, and $f \in \{R, \nn{R}{R}, \nn{S}{R}\}$.
We measured mean SSIM values of $0.50$ for $\nn{R}{R}$, $0.47$ for $\nn{S}{R}$, and $0.34$ for $R$.
This means that the influence of the label is smaller when gradients are propagated through the AE,
but also emphasizes how dissimilar gradients of just ResNet 50 are, compared to any AE at just $0.12$ to $0.18$ SSIM (\autoref{tab:gradientssim}).

\begin{figure}[b]
\includegraphics{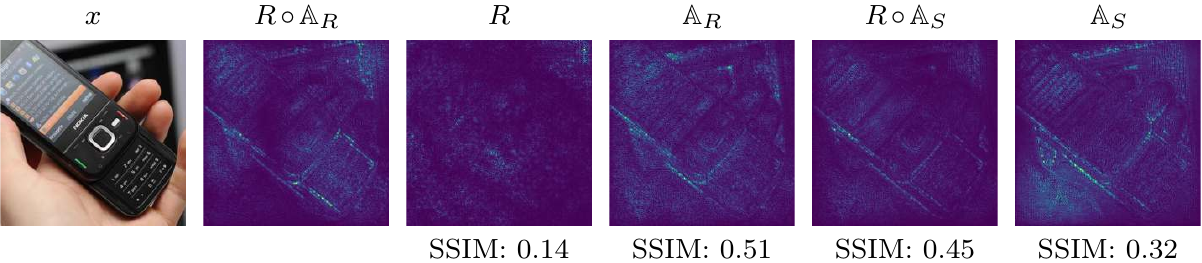}
\caption{Gradient magnitudes for ResNet 50 ($ R $) given a single input $x$, propagated through $\autoe{R}$ or $\autoe{S}$. SSIM values are in comparison to $||\nabla_x \nn{R}{R}(x)||.$}
\label{fig:gradientssim}
\end{figure}

\autoref{fig:gradientssim} visualizes this phenomenon.
Here, gradient magnitudes observed for just AEs ($\autoe{S}$, $\autoe{R}$) predominantly highlight edges as source of error.
This is expected, since it is more difficult to accurately reproduce the high frequencies required by sharp edges, compared to the lower frequencies of blobs.
Extracting magnitudes based on classification from AEs show similar structures ($\nn{R}{R}$, $\nn{S}{R}$).
Some coincidental overlap between ResNet 50 ($R$) and AE variants is unavoidable, since edges are also important for classification~\cite{zeiler2014visualizing}.
However, the classifier on its own differs considerably from all other patterns.
Overall, SSIM is at least twice as high between AE variants, than between the classifier and any of the AE configurations.

\section{Experiments}
\label{sec:experiments}

\begin{figure}[t]
	\scriptsize
	\includesvg[svgpath=figures/,width=\linewidth]{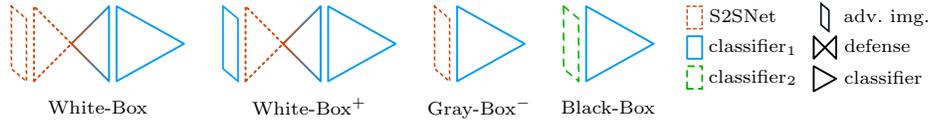}
	\caption{Overview of analyzed attack/defense scenarios.}
	\label{fig:overview}
\end{figure}

This section presents experiments quantifying the robustness of \netnames when used as a defense against adversarial attacks (Figure~\ref{fig:overview}).
The experimental setup closely follows the conditions from Guo~\etal~\cite{guo2018countering} in order to facilitate comparability:

\begin{itemize}
\item \textbf{Dataset}: We use ImageNet to test \netnames in a challenging, large-scale scenario.
Classifiers are trained on its training set and evaluations are carried out on the full validation set $X$ (50000 images).

\item \textbf{Image classifier under attack}: we use ResNet 50 ($ R $) and Inception v3 ($ I $), both pre-trained on ImageNet as target models.
The classifiers have been trained under clean conditions \ie, no special considerations with respect to adversarial attacks were made during training.

\item \textbf{Defense}: we train an \netname for each of the classifiers under attack, following the scheme described in \autoref{sec:normnet}.
These defenses are denoted as $ \autoe{R} $ and $ \autoe{I} $ for $ R $ and $ I $ respectively.

\item \textbf{Perturbation Magnitude}: we use the normalized $ L_2 $ norm (\ltwo) between a clean sample $ x $ and its adversary $ \hat{x} =  x+ \epsilon \cdot \delta $, as defined by Guo~\etal~\cite{guo2018countering}.
Epsilon values for each attack are listed below.

\item \textbf{Defense Strength Metric}: vulnerability to adversarial attacks is measured in terms of the number of newly misclassified samples.
More precisely, for any given attack to classifier $f$, we calculate
$\frac{\sum_{x \in TP} \mathcal{L}_2(x, \hat{x})}{|TP|}$, where $TP = \{ x \in X | f(x) = y^* \}$ is the set of true positives and $\hat{x}$ is the adversarial example generated by the attack, based on $x$.

\item \textbf{Attack Methods}: protected models are tested against a single step method, an iterative variant, and an optimization-based alternative.
Note that to replicate realistic threat conditions, all resulting adversarial samples are cast to the discrete RGB range [0, 255].
\begin{itemize}
  \item \textit{Fast Gradient Sign Method (FGSM)}~\cite{goodfellow2014explaining}: a simple, yet effective attack that works also when transferred to different models. Epsilon values used for this method are $\epsilon \in \{0.5, 1, 2, 4, 8, 16\}$.
  
  \item \textit{Basic Iterative Method (BIM)}~\cite{kurakin2016physical}: an iterative version of FGSM that shows more attack effectiveness but less transferable properties. Epsilon values used for BIM are $\epsilon \in \{0.5, 1, 2, 4, 8\}$. The number of iterations is fixed at $10$.
  
  \item \textit{Carlini-Wagner $L_2$ (CW)}~\cite{carlini2017towards}: an optimization-based method that has proven to be effective even against hardened models.
  Note that this attack issues perturbations that lay in a continuous domain.
  Epsilon values used for CW are $\epsilon \in \{0.5, 1, 2, 4\}$; the number of iterations is fixed at 100, $\kappa=0$ and $\lambda_f=10$.
\end{itemize}

\item \textbf{Attack Conditions}: we test the proposed defenses under two conditions.
\begin{itemize}
  \item \textit{White-box setting}: the attacker has knowledge about the classifier and the defense.
  This includes reading access to both the predictions of the classifier, intermediate activations and backpropagated gradients.
  The attacker is forced to forward valid images through the defended network \ie, through the composed classifier $ \nn{R}{R} $ or $ \nn{I}{I} $.
  
  \item \textit{Gray-box setting}: the attacker has access to a classification model but is unaware of its defense strategy.
\end{itemize}

\end{itemize}

\subsection{White-Box}
\label{sec:whitebox}

\begin{figure}[b]
	\centering
	\includegraphics{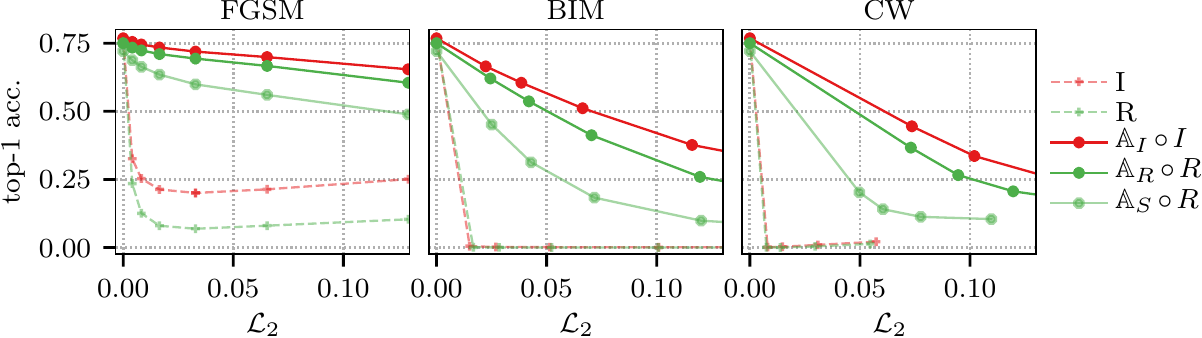}
	\caption{White-box attacks on ResNet 50 ($ R $) and Inception v3 ($ I $), with (solid) and without (dashed) \netname as a defense.}
	\label{fig:whitebox-resnet}
\end{figure}

For this setting, input images flow first through the \netname before reaching the original classifier.
Similarly, gradients are read from the shallowest layer of the \netname.
For comparison, attacks are also run on the unprotected versions of ResNet 50 and Inception v3.
Results are summarized in \autoref{fig:whitebox-resnet}.

Under these conditions, plain models are only able to moderately resist FGSM attacks, and fail completely for the more capable BIM and CW attacks.
In contrast, \netnames provide high levels of protection.
CW, being an optimization attack and generally the most capable in our comparison, is expected to be more effective at breaking through the \netname defense.
However, it does so by introducing large perturbations even for small values of $ \epsilon $.
In fact, none of the attacking configurations was able to entirely fool any of our defended classifiers for even the highest levels of \ltwo in our tests.
Note that these white-box results are, to our surprise, already comparable with some state-of-the-art gray-box defenses~\cite{guo2018countering,xie2018mitigating} (\ie, known classifier but unknown defense).
Despite the less favorable conditions for a white-box defense, \netnames already match alternative state-of-the-art protections that were tested under the more permissive assumptions allowed in gray-box settings.

For completeness, we also evaluate the defense using a pre-trained SegNet($\autoe{S}$) instead of an \netname for ResNet 50 (shown in \autoref{fig:whitebox-resnet} with light solid green).
This is most similar to the defense proposed by Meng~\etal~\cite{meng2017magnet}.
We can confirm that having just an AE does not suffice to guard the classifier against adversarial attacks and its performance is consistently lower than \netnames.
For a visual analysis of the attacks and their reconstructions by \netnames, we refer the reader to the supplementary material.

\subsubsection{Bypassing \netnames through Reparametrization}
In order to push the limits of \netnames, we now simulate a more hostile scenario where an attacker tries to actively circumvent the defense mechanism.
Based on the work of Athalye~\etal~\cite{athalye2018obfuscated}, a reparametrization of the input space can be implemented for defenses that operate under the composition of functions $ f \circ g $.
This works by defining the original input $ x $ as a function of a hidden state $ x = h(z) $ such that $ g(h(z)) = h(z) $.
Note that the main motivation behind reparametrization is to alleviate instabilities of gradients caused by defenses that \emph{rely} on said instability.
Although \netnames are very deep architectures, the resilience of this defense lays in the directed change that has been induced within the information contained in gradients.
In addition, there is no trivial way to come up with a suitable $ h(z) $ that does not end up inducing the same transformation in the gradients, as done by \netnames.

\begin{figure}[t]
	\centering
	\includegraphics{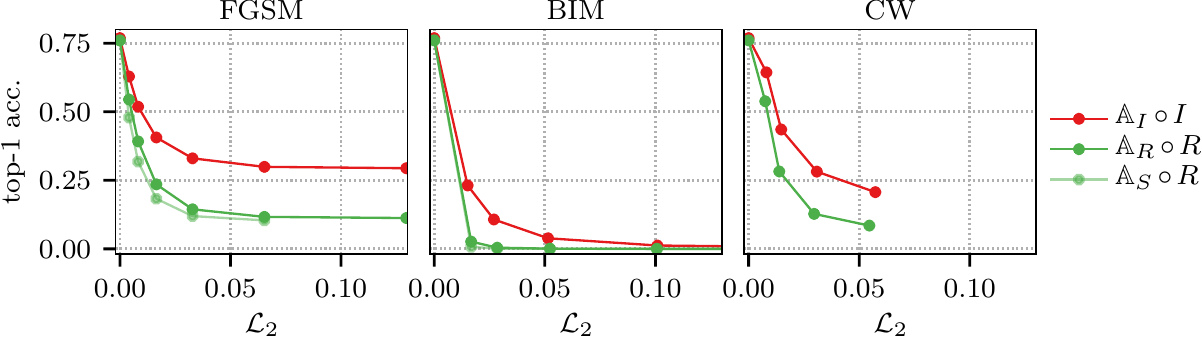}
	\caption{\wba attacks on ResNet 50 ($ R $) and Inception v3 ($ I $) classifiers.}
	\label{fig:grayboxplus}
\end{figure}

Despite all these concerns, we assume for this experiment that such $ h(z) $ can be found, and that attacking $ h(z) $ does indeed circumvent \netnames altogether.
We simulate the potential strength of this attack by using gradients of the original (unprotected) classifier, and applying them to the input directly.
The resulting adversarial attack is then passed through the hardened classifier.
We refer to this hypothetical scenario as \wba.
Results are shown in \autoref{fig:grayboxplus}.

Under these conditions, we observe that hardened models revert back to the behavior shown by their corresponding unprotected versions for FGSM and BIM.
In general, this condition is expected, and confirms once more that \netnames are being trained to preserve the information that is useful to the classifier.
Gradients collected directly from a vulnerable model, have by definition, only information that is useful for classification and hence, perturbations based on those gradients will be preserved by \netnames.
Interestingly, while CW is most successful in the previous white-box experiment, its highly optimized perturbations are less effective here.
We believe that CW is ``overfitting'' more strongly on the adversarial signal of the original model than what \netnames find useful to preserve.

\begin{figure}[b]
	\centering
	\includegraphics{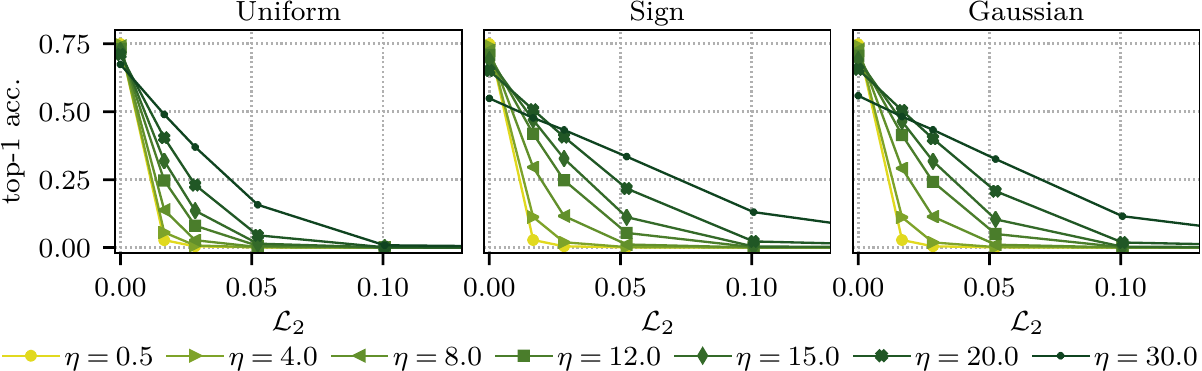}
	\caption{\wba attacks on $\nn{R}{R}$ with BIM, mitigated with additive sign, random and Gaussian noise of varying strength $\eta$.}
	\label{fig:grayboxplusnoise}
\end{figure}

At this point, we can further exploit the benefits that \netnames enjoy as a defense, by adding another layer of protection and showing how it affects the effectiveness of an attack.
As shown by Palacio~\etal~\cite{palaciofolz2018normnet}, AEs following a structure-to-signal training scheme exhibit strong resilience to random noise, as opposed to traditionally trained AEs.
We demonstrate that it is straightforward to add a layer of protection based on random noise.
Said stochastic strategy is added after the adversarial image has been computed but before it passes through an \netname defense.
We experiment with three sources of noise:
\begin{itemize}
	\item \textit{Gaussian Noise}: $ x \sim \frac{1}{\sqrt{2\pi \sigma^2}} e^{-(x^2/(2\sigma^2))}$, where $\sigma=\eta$.
	\item \textit{Uniform Noise}: $ \eta x $, where $x \sim \mathcal{U}(-1, 1)$.
	\item \textit{Sign Noise}: $ \eta \mathtt{sgn}(x)$, where $x \sim \mathcal{U}(-0.5, 0.5)$
\end{itemize}
\autoref{fig:grayboxplusnoise} shows the results of a BIM attack for ResNet 50, with noise levels $ \eta \in \{0.5, 4, 8, 12, 15, 20, 30\} $ under \wba conditions.
Overall, as noise strength increases, the resilience to adversarial attacks improves.
The initial degradation under zero adversarial attacks is dependent on the amount of noise $ \eta $ which can be tuned as the trade-off between maximum accuracy and adversarial robustness.
We observe that both Gaussian and sign noise perform almost identically, while uniform noise offers the least effective protection.

\subsection{Gray-Box}
\label{sec:graybox}

In contrast to white-box scenarios, gray-box attacks assume that there is limited access to information about the attack target.
More specifically, the conditions for gray-box define that the attacker has knowledge about the network but not about the defense strategy.

Given the compositional nature of \netnames, a novel way to defend a classification network under gray-box conditions consists in giving access to the gradients of \netnames.
Once an attack is ready, the \netname is \emph{removed} and the perturbed image is processed by the classifier only.
In other words, for a classifier $f$ and corresponding defense $g$, the attack is crafted based on the combined network $f \circ g$ but forwarded to just the classifier $f$.
We refer to this type of defense as \gm.

\begin{figure}[t]
	\centering
	\includegraphics{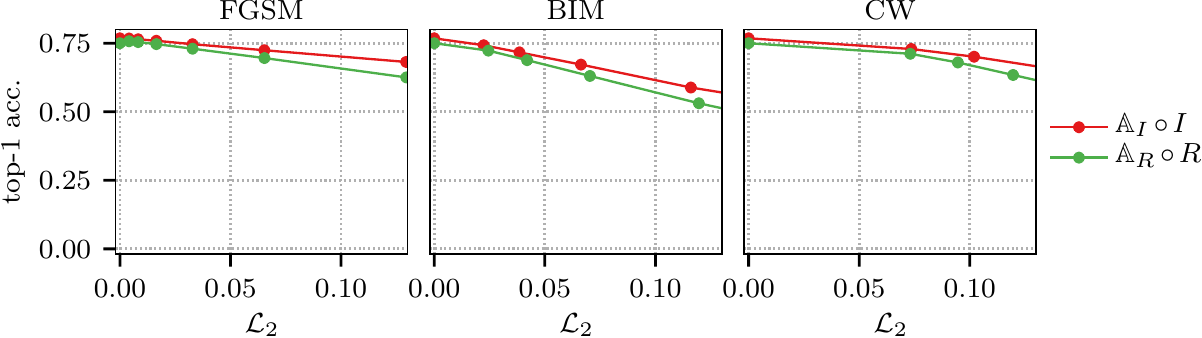}
	\caption{\gm attacks on ResNet 50 ($ R $) and Inception v3 ($ I $).}
	\label{fig:grayboxminus}
\end{figure}

We run the attacks with the same experimental settings described in \autoref{sec:whitebox}, but enforce a defense policy following \gm conditions.
Results are presented in \autoref{fig:grayboxminus}.
Overall, we see that this scenario is consistently robust to any adversarial attack.
Fooling perturbations are clearly visible and cannot be considered adversarial anymore.
With BIM-based attacks, classifiers gain back roughly half as much accuracy as in the white-box case.
Again, CW is more comparable to FGSM than BIM, i.e., mostly ineffective, and consistent with observations made in the \wba case.

By combining the results of \gm and \wba, we can reason about the relationship of $ \pert{x}{f} $ and $ \pert{x}{f\circ g} $.
First, the \wba experiment tells us that attacks crafted for $ f $ are valid attacks for $ f \circ g $.
Following from the definition of these spaces, it holds that $  \pert{x}{f} \subsetsim \pert{x}{f\circ g}  $.
Furthermore, looking at the \gm experiments, taking elements from $ \nabla_x f \circ g $ to create an attack, produces elements in $ \pert{x}{f \circ g} $.
As these do not attack $ f $, we conclude that those same elements do not lie in $ \pert{x}{f} $.
We finally conclude that with the tested attacks $ \pert{x}{f} $ cannot be reachable from $ \nabla_x f \circ g $.

\subsection{Classifier Relationships in Adversarial Space}
\label{sec:blackbox}

Given the high robustness of \netnames in the \gm scenario, we do not expect further insights from traditional black-box attacks.
Instead, we explore the relationships between the signal that different image classifiers use, as reported in~\cite{palaciofolz2018normnet}.
Intuitively, if a signal used by a classifier $f_1$ also encompasses the signal that another classifier $f_2$ uses, then an adversarial attack crafted for $f_1$ should also fool $f_2$.
Note that the relation is directional, and it may not hold in the opposite way.

\begin{figure}[t]
	\includegraphics[width=\textwidth]{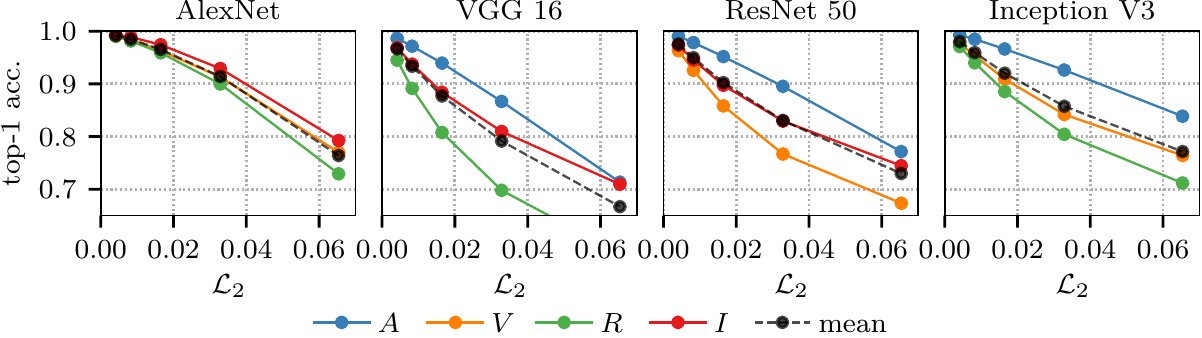}
	\caption{Relative accuracy for black-box attacks on AlexNet ($ A $), VGG 16 ($ V $), ResNet 50 ($ R $), and Inception v3 ($ I $).}
	\label{fig:blackbox}
\end{figure}

To test this, we construct adversarial samples using FGSM for four reference classifiers: AlexNet~\cite{DBLP:journals/corr/Krizhevsky14}, VGG 16~\cite{Simonyan14c}, ResNet 50, and Inception v3.
For each classifier we run black-box attacks with perturbations computed on the other three models.
The resulting accuracies (in terms of their relative drop in accuracy) are shown in \autoref{fig:blackbox}.

As expected, adversarial examples from higher compatible architectures result in higher fooling rations (lower accuracy).
Overall, adversarial samples created for AlexNet are the least compatible among the four, mainly due to its comparably lower accuracy and limited amount of input signal used.
On the other hand, ResNet 50 produces the most compatible perturbations, being the model generating attacks that were most effective when tried on other architectures.

Furthermore, we can also analyze how easy to fool a network is, based on the following criteria:
\begin{itemize}
  \item Fastest drop in accuracy per classifier.
  \item Mean accuracy of adversarial samples coming from other classifiers that have the same \ltwo.
\end{itemize}

With either criterion, results clearly show that VGG 16 is the easiest to fool, coming up on first place, followed by ResNet 50, AlexNet, and finally Inception v3.
This relationship in terms of useful signal, aligns with results in~\cite{palaciofolz2018normnet}.
Furthermore, similar experiments in~\cite{moosavi2017universal} based on universal perturbations, indicate that a selection of similar architectures (CaffeNet, VGG 19, ResNet 101 and GoogLeNet) maintain the same relationship.

\section{Conclusions \& Future Work}
\label{sec:discussion}

We have proposed \netnames as a new method to defend neural networks against adversarial examples.
We model the defense strategy as a transformation of the domain used by attackers, namely gradients coming from the attacked classifier.
Instead of focusing on gradient obfuscation via non-differentiable methods or any other instabilities, we purposely induce a transformation on the gradients that strip them from any semantic information.
\netnames work by masking the classifier via the function composition.
That way, the information of inputs is preserved for classification, but gradients point to structural changes for reconstructing the original sample.
Such a defense is possible by using a novel two-stage \emph{Structure-to-Signal} training scheme for deep AEs.
On the first stage, the AE is trained unsupervised in the traditional way.
For the second part, only the decoder gets fine-tuned with gradients from the model that is to be defended.

We evaluate the proposed defense under white-box and gray-box settings using a large scale dataset, against three different attack methods, on two highly performing deep image classifiers.
A baseline comparison shows that the two-staged training scheme performs better than using regular AEs.
Most interestingly, we show that resiliency to adversarial noise under white-box conditions, exhibit comparable performance to state-of-the-art under more favorable gray-box settings.
Furthermore, we show how the properties of \netnames can be exploited to add more defense mechanisms to maintain robustness even under the harshest, albeit currently hypothetical conditions, where the protection of \netnames is circumvented.
A gray-box scenario was also tested where the defense consists on the removal of \netnames, showing high levels of robustness for all attacks.
Finally, a comparison between the resiliency of four well-known deep CNNs is presented, providing further evidence that a relation of order exists between these classifiers, in terms of the amount of signal they use; this time, in terms of the effectiveness of adversarial noise.

\netnames are only one way in which the transformation can occur in gradient space.
We would like to explore other ways in which such transformation can occur, and even if the intersection between gradients yielding successful adversarial perturbations can be effectively zero.
The signal-preserving nature of \netnames make this defense a potential mechanism to explore and understand the nature of attacks.
Comparing classification consistency of a clean sample, before and after being passed through a \netname, has potential implications for detection of adversarial attacks by learning abnormal distribution fluctuations.

\section*{Acknowledgments} This work was supported by the BMBF project DeFuseNN (Grant 01IW17002) and the NVIDIA AI Lab (NVAIL) program.
We thank all members of the Deep Learning Competence Center at the DFKI for their comments and support.

\bibliographystyle{splncs}
\bibliography{resistnet}

\newpage

\appendix
\section{On the Relationship between $\pert{x}{f}$ and $\pert{x}{\fog}$: Formal Proof}

Experiments of \psec{sec:gradientnature} show that the domain of attacks is being shifted such that their range does not lie on the vulnerable space $\pert{x}{f}$.
We conduct a formal analysis of the relationship between perturbation spaces $\pert{x}{f}$ and $\pert{x}{\fog}$ and show that one is a subset of the other.
There are a few simplifications required for the proof, which will be accounted for at the end of this section.
First, we use the results of BIM for an \ltwo$ = 0.05$ tested on ResNet 50 as a reference.
Values for the baseline (no defense), \wba and, \gm are 0.0, 0.0 and 0.672041 respectively.
Computing the relative drop in performance (\textit{i.e.}, normalizing by 0.75004: the accuracy under no attack) yields 0.0, 0.0 and 0.8960.
To simplify the handling of fuzzy sets, we assign the membership functions $\mu(\alpha(x))$ for each perturbation set to be either 0 or 1 if the fooling ratio is below random chance or above 0.8 respectively.
With this in mind, we can define four axioms that come from the domain of the adversarial attack $\alpha: \nabla_f \cup \nabla_{\fog} \rightarrow \mathcal{P}$, and the aforementioned experiments with simplified membership functions.
The proof is a simple proof by contradiction with case analysis on the first disjunction.

\begin{proof}\leavevmode
    \begin{enumerate}
    \item $\forall_x(x\in \nabla_f \lor x\in \nabla_{\fog})$ \hfill(Def. domain of $\alpha$)
    \item $\forall_x(x\in \nabla_f \imp \alpha(x) \in \pert{x}{f})$ \hfill(Baseline Exp.)
    \item $\forall_x(x\in \nabla_f \imp \alpha(x) \in \pert{x}{\fog})$ \hfill(\wba)
    \item $\forall_x(x\in \nabla_{\fog} \imp \alpha(x) \not\in \pert{x}{f})$ \hfill(\gm)
    
    \hspace{\dimexpr\itemindent-1.3em}\rule{\dimexpr\linewidth-\itemindent+1.3em}{1pt}
    
    \begin{level}
    \item $\exists_x(\alpha(x) \in \pert{x}{f} \land \alpha(x) \not\in \pert{x}{\fog})$ \hfill (Assumption, $\not\subseteq$)
    \item $\alpha(a) \in \pert{a}{f} \land \alpha(a) \not\in \pert{a}{\fog}$ \hfill (Skolemization $x \rightarrow a, 5$)
    
    \item $a\in \nabla_f \lor a\in \nabla_{\fog}$ \hfill (U.I. $x \rightarrow a$, 1)
    \item $a \in \nabla_f \imp \alpha(a) \in \pert{a}{\fog}$ \hfill (U.I. $x \rightarrow a$, 3)
    \item $a \in \nabla_{\fog} \imp \alpha(a) \not\in \pert{a}{f}$ \hfill (U.I. $x \rightarrow a$, 4)
    
    \begin{level}
    
        \hspace{\dimexpr\itemindent-1.7em}\rule{\dimexpr\linewidth-\itemindent+1.7em}{1pt}
        
        \item $a \in \nabla_{\fog}$ \hfill (Assumption, 7)
        
        \item $\alpha(a) \not\in \pert{a}{f}$ \hfill ($\imp$ 10, 9)
        \item $\alpha(a) \not\in \pert{a}{f} \land \alpha(a) \in \pert{a}{f}$ \hfill ($\land$, 11, 6)
        \item Contradiction! \hfill \Lightning
    
        \hspace{\dimexpr\itemindent-1.7em}\rule{\dimexpr\linewidth-\itemindent+1.7em}{1pt}
        
    \end{level}
    \item $\lnot(a \in \nabla_{\fog})$ \hfill (Q.E.A. 10)

    \item $a \in \nabla_{f}$ \hfill (D.Syllogism, 14, 7)
    \item $\alpha(a) \in \pert{a}{\fog}$ \hfill ($\imp$ 15, 8)
    \item $\alpha(a) \not\in \pert{a}{\fog} \land \alpha(a) \in \pert{a}{\fog}$ \hfill ($\land$, 6, 16)
    \item Contradiction! \hfill \Lightning

    \hspace{\dimexpr\itemindent-1.7em}\rule{\dimexpr\linewidth-\itemindent+1.7em}{1pt}

    \end{level}
    
    \item $\lnot\exists_x(\alpha(x) \in \pert{x}{f} \land \alpha(x) \not\in \pert{x}{\fog})$ \hfill (Q.E.A., 5)
    \item $\forall_x\lnot(\alpha(x) \in \pert{x}{f} \land \alpha(x) \not\in \pert{x}{\fog})$ \hfill ($\lnot\exists$, 19)
    \item $\forall_x(\alpha(x) \not\in \pert{x}{f} \lor \alpha(x) \in \pert{x}{\fog})$ \hfill (Distr. $\lnot$, 20)
    \item $\forall_x(\alpha(x) \in \pert{x}{f} \imp \alpha(x) \in \pert{x}{\fog})$ \hfill (MI $\imp$., 21)

    \item $\pert{x}{f} \subseteq \pert{x}{\fog}$ \hfill (Def. $\subseteq$, 22)
    
    \end{enumerate}
\end{proof}

Going back to the simplified membership function, it follows that different reference experiments (network, attack and \ltwo) will naturally yield different results.
This is especially true if we take the raw accuracy as the membership function, instead of the simplified one.
However, one can argue that the subset relationship is, in general terms, valid since overall, most experiments under the simplified membership function yield the same axioms).
This is why we denote such a relationship by the approximate subset relationship $\subseteq$ to refer to this result.

\begin{figure}
\centering
\def\svgwidth{\columnwidth}
\scriptsize
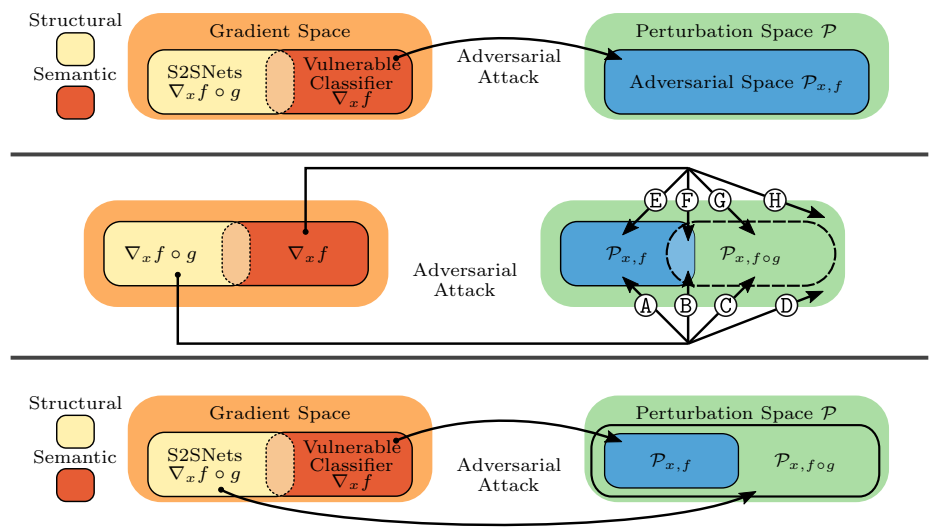
\caption{Analysis of the relationship between gradient space and perturbation space. By doing case analysis (middle) and a proof by contradiction, we infer that the perturbation space approximates $\pert{x}{f} \protect\subsetsim \pert{x}{\fog}$ }
\label{fig:proof}
\end{figure}

A similar analysis can be done graphically as shown in Figure \ref{fig:proof} (middle).
Case $\mathtt{A}$ is covered by \gm experiments which show that $\pert{x}{f}$ can only be reached 10.4\% of the time, which we know now, actually corresponds to case $\mathtt{B}$ due to the inclusion $\pert{x}{f} \subsetsim \pert{x}{\fog}$.
This leaves cases $\mathtt{C,D}$ which are covered by \wb experiments.
Here, using the BIM on ResNet 50 and \ltwo=~0.05 as reference, yields that $\alpha(x) \in \pert{x}{\fog}$ with a membership $\mu(\alpha(x)) = 1-(0.5/0.75) = 0.333$.
That leaves case $\mathtt{D}$ with perturbations in the remaining 0.667 in $\mathcal{P} - \{\pert{x}{f} \cup \pert{x}{\fog}\}$.

Likewise, for cases where the domain is $\nabla_f$, we see that they all fall into $\pert{x}{f}$ which we know lies in $\pert{x}{\fog}$ hence, they all fall into case $\mathtt{F}$, eliminating samples falling into the remaining ones $\mathtt{E, G, H}$.
\section{Perturbed Images and their Reconstructions}
The following figures (\ref{fig:attack0}-\ref{fig:attack6}) show attack attempts on ResNet 50 ($R$) with correctly classified images from our random shuffle of the ImageNet validation set.
$x$ (column 1) denotes the clean image and $\hat{x}$ its perturbed variants.
Columns 2-7 show a combination of attack method (FGSM, BIM, CW) and gradient sources ($\nabla_x R(x)$, $\nabla_x \nn{A}{R}(x)$).
Below are perturbations $\delta$ (row 2), reconstructions $\hat{x}' = \autoe{R}(\hat{x})$ (row 3), and remaining perturbation in reconstructions $\delta' = \autoe{R}(\hat{x}) - \autoe{R}(x)$ (row 4).
Perturbation images are subject to histogram equalization for increases visibility.
These examples further illustrate the structural nature of attacks through S2SNets.

\begin{figure}[ht]
\centering
\includegraphics{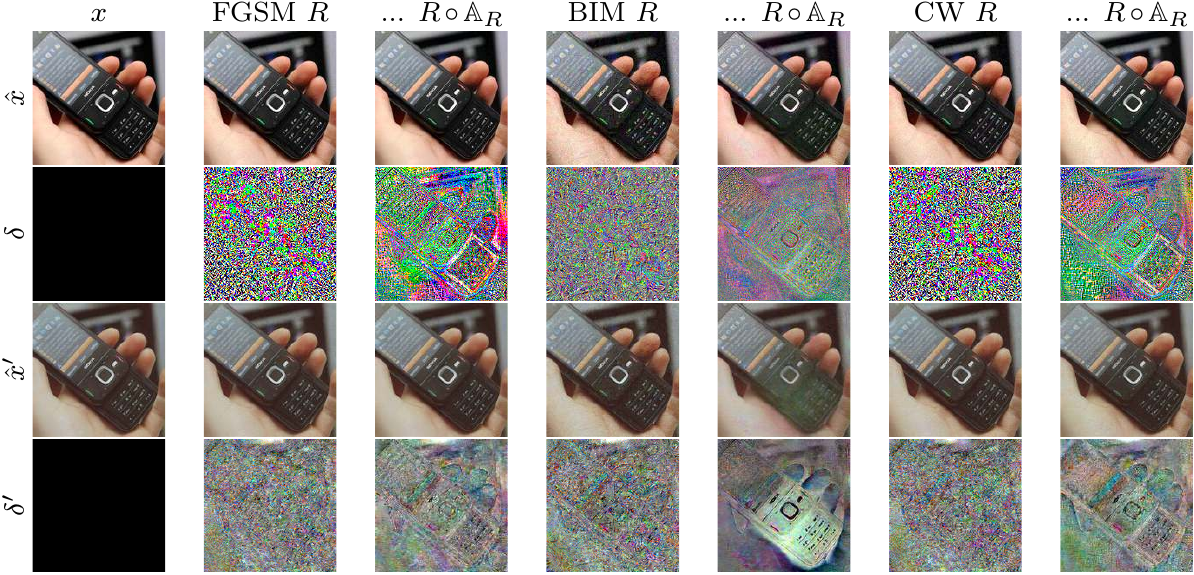}
\caption{Example of different attacks on ResNet 50 ($R$) with and without $\autoe{R}$, along with perturbations and reconstructions. \ltwo (left to right): $0.032$, $0.032$, $0.098$, $0.115$, $0.052$, $0.064$.}
\label{fig:attack0}
\end{figure}

\begin{figure}[ht]
\centering
\includegraphics{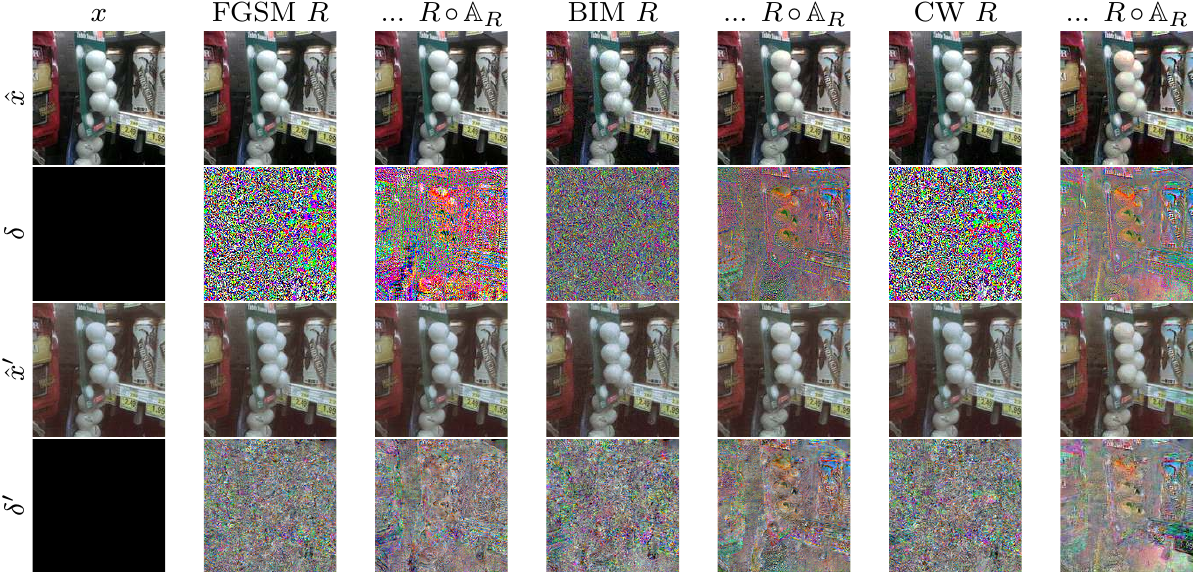}
\caption{Example of different attacks on ResNet 50 ($R$) with and without $\autoe{R}$, along with perturbations and reconstructions. \ltwo (left to right): $0.036$, $0.036$, $0.111$, $0.104$, $0.059$, $0.172$.}
\label{fig:attack1}
\end{figure}

\begin{figure}[ht]
\centering
\includegraphics{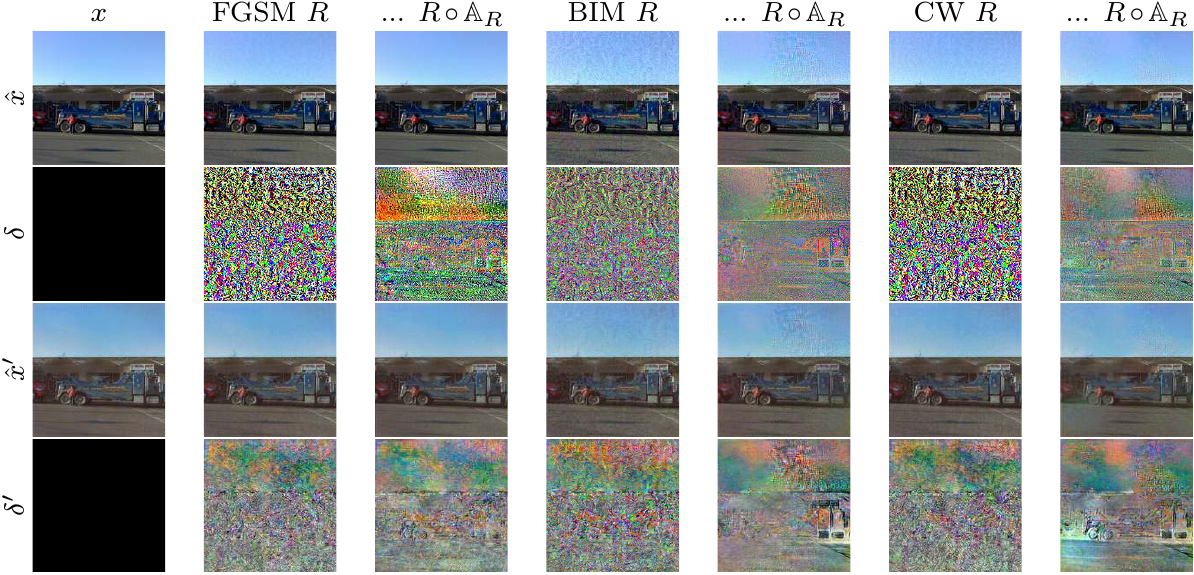}
\caption{Example of different attacks on ResNet 50 ($R$) with and without $\autoe{R}$, along with perturbations and reconstructions. \ltwo (left to right): $0.028$, $0.028$, $0.084$, $0.105$, $0.045$, $0.095$.}
\label{fig:attack2}
\end{figure}

\begin{figure}[ht]
\centering
\includegraphics{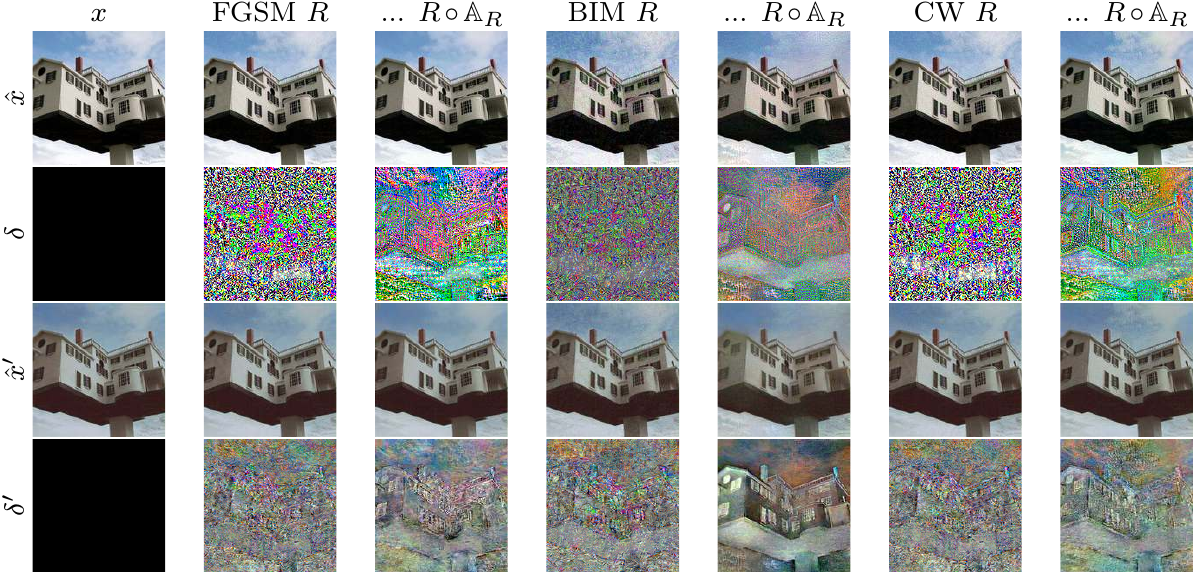}
\caption{Example of different attacks on ResNet 50 ($R$) with and without $\autoe{R}$, along with perturbations and reconstructions. \ltwo (left to right): $0.028$, $0.028$, $0.088$, $0.119$, $0.046$, $0.054$.}
\label{fig:attack3}
\end{figure}

\begin{figure}[ht]
\centering
\includegraphics{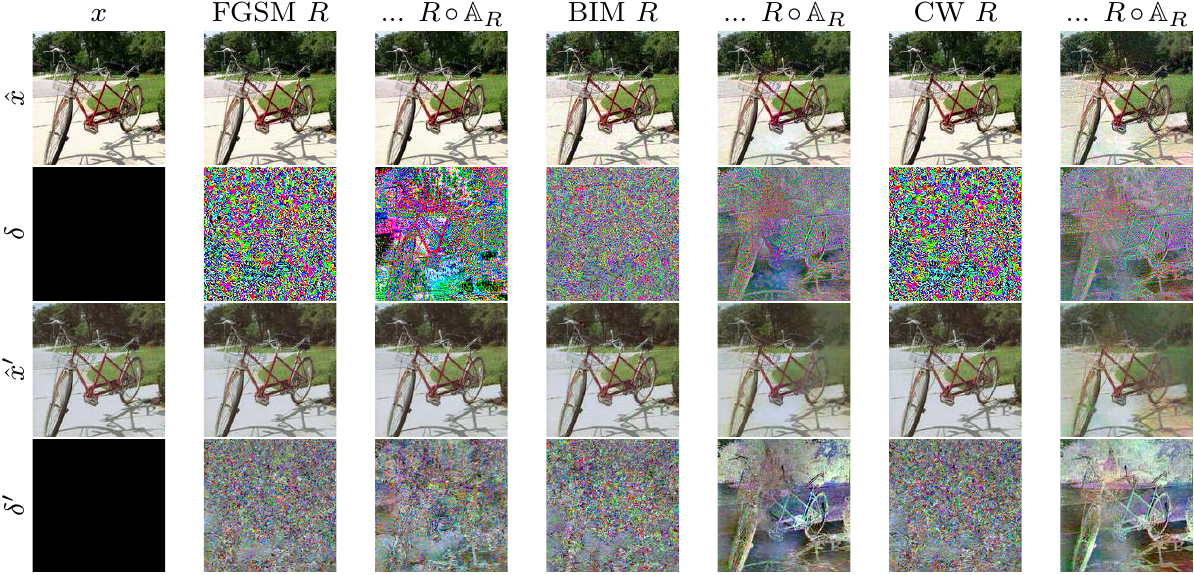}
\caption{Example of different attacks on ResNet 50 ($R$) with and without $\autoe{R}$, along with perturbations and reconstructions. \ltwo (left to right): $0.025$, $0.025$, $0.077$, $0.124$, $0.041$, $0.212$.}
\label{fig:attack4}
\end{figure}

\begin{figure}[ht]
\centering
\includegraphics{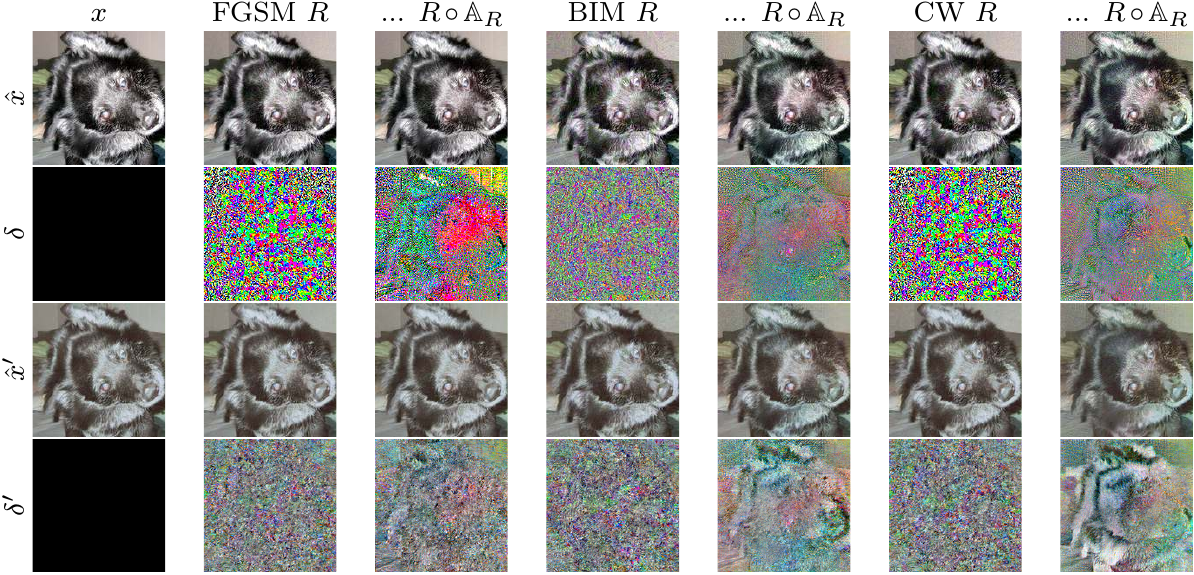}
\caption{Example of different attacks on ResNet 50 ($R$) with and without $\autoe{R}$, along with perturbations and reconstructions. \ltwo (left to right): $0.030$, $0.030$, $0.095$, $0.107$, $0.050$, $0.150$.}
\label{fig:attack5}
\end{figure}

\begin{figure}[ht]
\centering
\includegraphics{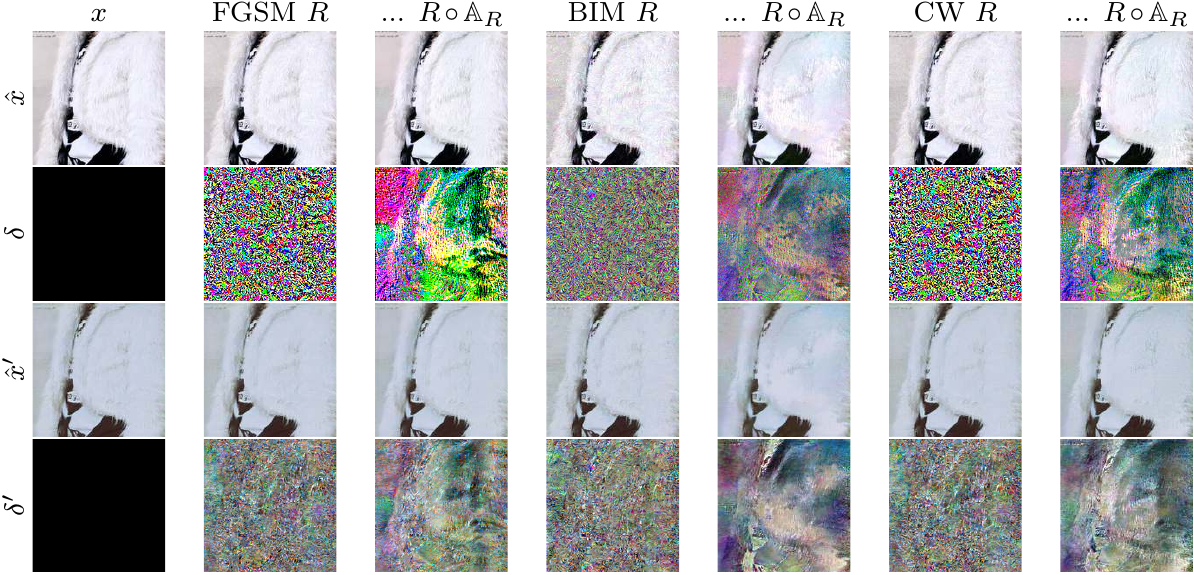}
\caption{Example of different attacks on ResNet 50 ($R$) with and without $\autoe{R}$, along with perturbations and reconstructions. \ltwo (left to right): $0.018$, $0.018$, $0.053$, $0.067$, $0.029$, $0.045$.}
\label{fig:attack6}
\end{figure}

\clearpage
\section{Raw Experiment Data}

This section contains raw values for all experiments we conducted.
A subset was used to create the plots seen in the paper, but we also provide results we ended up not using for our argumentation.
In the following tables, \ltwo denotes the normalized $L2$ dissimilarity (\autoref{eq:nl2}) wrt. reference image $x_1$ and another image $x_2$.
We also provide $L_\infty$ values, the usual infinity norm divided by $255$ (\autoref{eq:linf}).
Settings for attacks besides $\epsilon$ are defined in \psec{sec:experiments}.

\begin{equation}
    \mathcal{L}_2(x_1, x_2) = \frac{|| x_1 - x_2 ||_2}{|| x_1 ||_2}
    \label{eq:nl2}
\end{equation}

\begin{equation}
    L_\infty(x_1, x_2) = \frac{max(|x_1-x_2|)}{255}
    \label{eq:linf}
\end{equation}

Tables \ref{tab:wb-fgs}, \ref{tab:wb-ifgs}, and \ref{tab:wb-cw} contain \wb results for the respective attack method.
They cover experiments described in \psec{sec:whitebox}
Tables \ref{tab:gb-fgs}, \ref{tab:gb-ifgs}, and \ref{tab:gb-cw} contain results attacks where the network that supplied the gradients (source network) that the attack is based on is different from the network that is attacked (target network).
They cover \wba and \gm experiments described in \psec{sec:whitebox} and \ref{sec:graybox}.
Finally, \autoref{tab:noise-ifgs} contains results on BIM attacks mitigated by different types and strengths ($\eta$) of noise.

\begin{longtable}[]{cccccc}
\caption{Accuracy and distance statistics for FGSM attacks.} 
\label{tab:wb-fgs} 
\\
\toprule
Network & $\epsilon$ & top-1 acc. (\%) & \dots (adversarial) & \ltwo & $L_\infty$ \\
\midrule
\endhead
$A$ & $0.5$ & $56.48$ & $20.30$ & $0.00413$ & $0.00196$ \\
$A$ & $1.0$ & $56.48$ & $8.60$ & $0.00825$ & $0.00392$ \\
$A$ & $2.0$ & $56.48$ & $2.80$ & $0.01647$ & $0.00784$ \\
$A$ & $4.0$ & $56.48$ & $1.14$ & $0.03283$ & $0.01569$ \\
$A$ & $8.0$ & $56.48$ & $0.83$ & $0.06533$ & $0.03137$ \\
$A$ & $16.0$ & $56.48$ & $0.69$ & $0.12945$ & $0.06275$ \\
$I$ & $0.5$ & $76.86$ & $32.58$ & $0.00413$ & $0.00196$ \\
$I$ & $1.0$ & $76.86$ & $25.32$ & $0.00825$ & $0.00392$ \\
$I$ & $2.0$ & $76.86$ & $21.32$ & $0.01647$ & $0.00784$ \\
$I$ & $4.0$ & $76.86$ & $20.04$ & $0.03284$ & $0.01569$ \\
$I$ & $4.0$ & $76.86$ & $20.04$ & $0.03284$ & $0.01569$ \\
$I$ & $8.0$ & $76.86$ & $21.37$ & $0.06535$ & $0.03137$ \\
$I$ & $16.0$ & $76.86$ & $25.09$ & $0.12948$ & $0.06275$ \\
$R$ & $0.5$ & $76.03$ & $23.49$ & $0.00413$ & $0.00196$ \\
$R$ & $1.0$ & $76.03$ & $12.58$ & $0.00826$ & $0.00392$ \\
$R$ & $2.0$ & $76.03$ & $8.00$ & $0.01648$ & $0.00784$ \\
$R$ & $4.0$ & $76.03$ & $6.90$ & $0.03284$ & $0.01569$ \\
$R$ & $8.0$ & $76.03$ & $8.02$ & $0.06535$ & $0.03137$ \\
$R$ & $16.0$ & $76.03$ & $10.37$ & $0.12947$ & $0.06275$ \\
$V$ & $0.5$ & $73.28$ & $8.82$ & $0.00413$ & $0.00196$ \\
$V$ & $1.0$ & $73.28$ & $3.75$ & $0.00826$ & $0.00392$ \\
$V$ & $2.0$ & $73.28$ & $2.74$ & $0.01648$ & $0.00784$ \\
$V$ & $4.0$ & $73.28$ & $2.85$ & $0.03285$ & $0.01569$ \\
$V$ & $8.0$ & $73.28$ & $3.79$ & $0.06537$ & $0.03137$ \\
$V$ & $16.0$ & $73.28$ & $6.64$ & $0.12951$ & $0.06275$ \\
$A \circ\mathbb{A}_{A}$ & $0.5$ & $56.38$ & $54.90$ & $0.00413$ & $0.00196$ \\
$A \circ\mathbb{A}_{A}$ & $1.0$ & $56.38$ & $53.79$ & $0.00827$ & $0.00392$ \\
$A \circ\mathbb{A}_{A}$ & $2.0$ & $56.38$ & $52.37$ & $0.01650$ & $0.00784$ \\
$A \circ\mathbb{A}_{A}$ & $4.0$ & $56.38$ & $50.68$ & $0.03289$ & $0.01569$ \\
$A \circ\mathbb{A}_{A}$ & $8.0$ & $56.38$ & $48.56$ & $0.06545$ & $0.03137$ \\
$A \circ\mathbb{A}_{A}$ & $16.0$ & $56.38$ & $42.86$ & $0.12970$ & $0.06275$ \\
$I \circ\mathbb{A}_{I}$ & $0.5$ & $76.79$ & $75.46$ & $0.00413$ & $0.00196$ \\
$I \circ\mathbb{A}_{I}$ & $1.0$ & $76.79$ & $74.51$ & $0.00825$ & $0.00392$ \\
$I \circ\mathbb{A}_{I}$ & $2.0$ & $76.79$ & $73.41$ & $0.01647$ & $0.00784$ \\
$I \circ\mathbb{A}_{I}$ & $4.0$ & $76.79$ & $71.91$ & $0.03283$ & $0.01569$ \\
$I \circ\mathbb{A}_{I}$ & $8.0$ & $76.79$ & $69.89$ & $0.06532$ & $0.03137$ \\
$I \circ\mathbb{A}_{I}$ & $16.0$ & $76.79$ & $65.40$ & $0.12942$ & $0.06275$ \\
$R \circ\mathbb{A}_{R}$ & $0.5$ & $75.00$ & $73.35$ & $0.00413$ & $0.00196$ \\
$R \circ\mathbb{A}_{R}$ & $1.0$ & $75.00$ & $72.33$ & $0.00826$ & $0.00392$ \\
$R \circ\mathbb{A}_{R}$ & $2.0$ & $75.00$ & $70.99$ & $0.01648$ & $0.00784$ \\
$R \circ\mathbb{A}_{R}$ & $4.0$ & $75.00$ & $69.33$ & $0.03285$ & $0.01569$ \\
$R \circ\mathbb{A}_{R}$ & $8.0$ & $75.00$ & $66.65$ & $0.06535$ & $0.03137$ \\
$R \circ\mathbb{A}_{R}$ & $16.0$ & $75.00$ & $60.47$ & $0.12948$ & $0.06275$ \\
$V \circ\mathbb{A}_{V}$ & $0.5$ & $68.14$ & $65.20$ & $0.00413$ & $0.00196$ \\
$V \circ\mathbb{A}_{V}$ & $1.0$ & $68.14$ & $63.38$ & $0.00827$ & $0.00392$ \\
$V \circ\mathbb{A}_{V}$ & $2.0$ & $68.14$ & $61.01$ & $0.01650$ & $0.00784$ \\
$V \circ\mathbb{A}_{V}$ & $4.0$ & $68.14$ & $58.16$ & $0.03289$ & $0.01569$ \\
$V \circ\mathbb{A}_{V}$ & $8.0$ & $68.14$ & $54.85$ & $0.06544$ & $0.03137$ \\
$V \circ\mathbb{A}_{V}$ & $16.0$ & $68.14$ & $47.38$ & $0.12965$ & $0.06275$ \\
$I \circ\mathbb{A}_{S}$ & $0.5$ & $73.80$ & $71.37$ & $0.00413$ & $0.00196$ \\
$I \circ\mathbb{A}_{S}$ & $1.0$ & $73.80$ & $69.71$ & $0.00825$ & $0.00392$ \\
$I \circ\mathbb{A}_{S}$ & $2.0$ & $73.80$ & $67.63$ & $0.01647$ & $0.00784$ \\
$I \circ\mathbb{A}_{S}$ & $4.0$ & $73.80$ & $65.41$ & $0.03283$ & $0.01569$ \\
$I \circ\mathbb{A}_{S}$ & $8.0$ & $73.80$ & $63.10$ & $0.06532$ & $0.03137$ \\
$I \circ\mathbb{A}_{S}$ & $16.0$ & $73.80$ & $59.38$ & $0.12943$ & $0.06275$ \\
$R \circ\mathbb{A}_{S}$ & $0.5$ & $72.07$ & $68.75$ & $0.00413$ & $0.00196$ \\
$R \circ\mathbb{A}_{S}$ & $1.0$ & $72.07$ & $66.33$ & $0.00825$ & $0.00392$ \\
$R \circ\mathbb{A}_{S}$ & $2.0$ & $72.07$ & $63.47$ & $0.01647$ & $0.00784$ \\
$R \circ\mathbb{A}_{S}$ & $4.0$ & $72.07$ & $59.89$ & $0.03283$ & $0.01569$ \\
$R \circ\mathbb{A}_{S}$ & $8.0$ & $72.07$ & $56.03$ & $0.06533$ & $0.03137$ \\
$R \circ\mathbb{A}_{S}$ & $16.0$ & $72.07$ & $48.87$ & $0.12943$ & $0.06275$ \\
\bottomrule
\end{longtable}
\begin{longtable}[]{cccccc}
\caption{Accuracy and distance statistics for BIM attacks.} 
\label{tab:wb-ifgs} 
\\
\toprule
Network & $\epsilon$ & top-1 acc. (\%) & \dots (adversarial) & \ltwo & $L_\infty$ \\
\midrule
\endhead
$A$ & $0.5$ & $56.48$ & $0.11$ & $0.02278$ & $0.01961$ \\
$A$ & $1.0$ & $56.48$ & $0.05$ & $0.03842$ & $0.03922$ \\
$A$ & $2.0$ & $56.48$ & $0.03$ & $0.06572$ & $0.07843$ \\
$A$ & $4.0$ & $56.48$ & $0.02$ & $0.11770$ & $0.15686$ \\
$A$ & $8.0$ & $56.48$ & $0.01$ & $0.22130$ & $0.31373$ \\
$I$ & $0.5$ & $76.86$ & $0.50$ & $0.01508$ & $0.01961$ \\
$I$ & $1.0$ & $76.86$ & $0.20$ & $0.02698$ & $0.03921$ \\
$I$ & $2.0$ & $76.86$ & $0.15$ & $0.05147$ & $0.07841$ \\
$I$ & $4.0$ & $76.86$ & $0.10$ & $0.10090$ & $0.15682$ \\
$I$ & $8.0$ & $76.86$ & $0.08$ & $0.19724$ & $0.31366$ \\
$R$ & $0.5$ & $76.03$ & $0.10$ & $0.01669$ & $0.01961$ \\
$R$ & $1.0$ & $76.03$ & $0.05$ & $0.02847$ & $0.03922$ \\
$R$ & $2.0$ & $76.03$ & $0.02$ & $0.05232$ & $0.07843$ \\
$R$ & $4.0$ & $76.03$ & $0.01$ & $0.10067$ & $0.15686$ \\
$R$ & $8.0$ & $76.03$ & $0.00$ & $0.19510$ & $0.31373$ \\
$V$ & $0.5$ & $73.28$ & $0.34$ & $0.01724$ & $0.01961$ \\
$V$ & $1.0$ & $73.28$ & $0.27$ & $0.02920$ & $0.03922$ \\
$V$ & $2.0$ & $73.28$ & $0.23$ & $0.05305$ & $0.07843$ \\
$V$ & $4.0$ & $73.28$ & $0.18$ & $0.10117$ & $0.15686$ \\
$V$ & $8.0$ & $73.28$ & $0.12$ & $0.19571$ & $0.31372$ \\
$A \circ\mathbb{A}_{A}$ & $0.5$ & $56.38$ & $44.11$ & $0.02637$ & $0.01961$ \\
$A \circ\mathbb{A}_{A}$ & $1.0$ & $56.38$ & $35.57$ & $0.04537$ & $0.03922$ \\
$A \circ\mathbb{A}_{A}$ & $2.0$ & $56.38$ & $24.47$ & $0.07559$ & $0.07843$ \\
$A \circ\mathbb{A}_{A}$ & $4.0$ & $56.38$ & $12.69$ & $0.12549$ & $0.15686$ \\
$A \circ\mathbb{A}_{A}$ & $8.0$ & $56.38$ & $4.00$ & $0.22330$ & $0.31373$ \\
$I \circ\mathbb{A}_{I}$ & $0.5$ & $76.79$ & $66.48$ & $0.02235$ & $0.01961$ \\
$I \circ\mathbb{A}_{I}$ & $1.0$ & $76.79$ & $60.47$ & $0.03852$ & $0.03922$ \\
$I \circ\mathbb{A}_{I}$ & $2.0$ & $76.79$ & $51.12$ & $0.06634$ & $0.07843$ \\
$I \circ\mathbb{A}_{I}$ & $4.0$ & $76.79$ & $37.65$ & $0.11604$ & $0.15686$ \\
$I \circ\mathbb{A}_{I}$ & $8.0$ & $76.79$ & $23.03$ & $0.21053$ & $0.31373$ \\
$R \circ\mathbb{A}_{R}$ & $0.5$ & $75.00$ & $62.05$ & $0.02445$ & $0.01961$ \\
$R \circ\mathbb{A}_{R}$ & $1.0$ & $75.00$ & $53.64$ & $0.04197$ & $0.03922$ \\
$R \circ\mathbb{A}_{R}$ & $2.0$ & $75.00$ & $41.21$ & $0.07040$ & $0.07843$ \\
$R \circ\mathbb{A}_{R}$ & $4.0$ & $75.00$ & $25.93$ & $0.11964$ & $0.15686$ \\
$R \circ\mathbb{A}_{R}$ & $8.0$ & $75.00$ & $11.78$ & $0.21596$ & $0.31373$ \\
$V \circ\mathbb{A}_{V}$ & $0.5$ & $68.14$ & $46.16$ & $0.02467$ & $0.01961$ \\
$V \circ\mathbb{A}_{V}$ & $1.0$ & $68.14$ & $34.45$ & $0.04230$ & $0.03922$ \\
$V \circ\mathbb{A}_{V}$ & $2.0$ & $68.14$ & $22.14$ & $0.07086$ & $0.07843$ \\
$V \circ\mathbb{A}_{V}$ & $4.0$ & $68.14$ & $12.65$ & $0.11932$ & $0.15686$ \\
$V \circ\mathbb{A}_{V}$ & $8.0$ & $68.14$ & $6.33$ & $0.21275$ & $0.31373$ \\
$R \circ\mathbb{A}_{S}$ & $0.5$ & $72.07$ & $45.11$ & $0.02507$ & $0.01961$ \\
$R \circ\mathbb{A}_{S}$ & $1.0$ & $72.07$ & $31.29$ & $0.04294$ & $0.03922$ \\
$R \circ\mathbb{A}_{S}$ & $2.0$ & $72.07$ & $18.32$ & $0.07170$ & $0.07843$ \\
$R \circ\mathbb{A}_{S}$ & $4.0$ & $72.07$ & $9.86$ & $0.12014$ & $0.15686$ \\
$R \circ\mathbb{A}_{S}$ & $8.0$ & $72.07$ & $5.42$ & $0.21249$ & $0.31373$ \\
$R \circ\mathbb{A}_{S}$ & $16.0$ & $72.07$ & $1.09$ & $0.40976$ & $0.62744$ \\
\bottomrule
\end{longtable}
\begin{longtable}[]{cccccc}
\caption{Accuracy and distance statistics for CW attacks.} 
\label{tab:wb-cw} 
\\
\toprule
Network & $\epsilon$ & top-1 acc. (\%) & \dots (adversarial) & \ltwo & $L_\infty$ \\
\midrule
\endhead
$I$ & $0.5$ & $76.86$ & $0.16$ & $0.00805$ & $0.00587$ \\
$I$ & $1.0$ & $76.86$ & $0.29$ & $0.01475$ & $0.00966$ \\
$I$ & $2.0$ & $76.86$ & $0.99$ & $0.03084$ & $0.01848$ \\
$I$ & $4.0$ & $76.86$ & $2.13$ & $0.05731$ & $0.03350$ \\
$R$ & $0.5$ & $76.03$ & $0.10$ & $0.00752$ & $0.00511$ \\
$R$ & $1.0$ & $76.03$ & $0.02$ & $0.01387$ & $0.00850$ \\
$R$ & $2.0$ & $76.03$ & $0.42$ & $0.02964$ & $0.01644$ \\
$R$ & $4.0$ & $76.03$ & $1.32$ & $0.05459$ & $0.02917$ \\
$I \circ\mathbb{A}_{I}$ & $0.5$ & $76.80$ & $44.49$ & $0.07357$ & $0.11869$ \\
$I \circ\mathbb{A}_{I}$ & $1.0$ & $76.80$ & $33.58$ & $0.10195$ & $0.17756$ \\
$I \circ\mathbb{A}_{I}$ & $2.0$ & $76.80$ & $25.19$ & $0.13851$ & $0.24872$ \\
$I \circ\mathbb{A}_{I}$ & $4.0$ & $76.80$ & $19.72$ & $0.18622$ & $0.31932$ \\
$R \circ\mathbb{A}_{R}$ & $0.5$ & $75.01$ & $36.66$ & $0.07310$ & $0.11166$ \\
$R \circ\mathbb{A}_{R}$ & $1.0$ & $75.01$ & $26.59$ & $0.09461$ & $0.15410$ \\
$R \circ\mathbb{A}_{R}$ & $2.0$ & $75.01$ & $20.63$ & $0.11958$ & $0.19562$ \\
$R \circ\mathbb{A}_{R}$ & $4.0$ & $75.01$ & $16.68$ & $0.15458$ & $0.23814$ \\
$R \circ\mathbb{A}_{S}$ & $0.5$ & $72.07$ & $20.26$ & $0.04971$ & $0.06407$ \\
$R \circ\mathbb{A}_{S}$ & $1.0$ & $72.07$ & $14.08$ & $0.06044$ & $0.07838$ \\
$R \circ\mathbb{A}_{S}$ & $2.0$ & $72.07$ & $11.27$ & $0.07758$ & $0.09839$ \\
$R \circ\mathbb{A}_{S}$ & $4.0$ & $72.07$ & $10.45$ & $0.10965$ & $0.13377$ \\
\bottomrule
\end{longtable}
\begin{longtable}[]{ccccc}
\caption{Accuracy for FGSM attacks when source and target network differ.} 
\label{tab:gb-fgs} 
\\
\toprule
Source Network & Target Network & $\epsilon$ & top-1 acc. (\%) & \dots (adversarial) \\
\midrule
\endhead
$A$ & $I$ & $0.5$ & $76.86$ & $56.11$ \\
$A$ & $R$ & $0.5$ & $76.03$ & $55.93$ \\
$A$ & $V$ & $0.5$ & $73.28$ & $56.01$ \\
$A$ & $A \circ\mathbb{A}_{A}$ & $0.5$ & $56.38$ & $56.26$ \\
$A$ & $I$ & $1.0$ & $76.86$ & $55.86$ \\
$A$ & $R$ & $1.0$ & $76.03$ & $55.43$ \\
$A$ & $V$ & $1.0$ & $73.28$ & $55.58$ \\
$A$ & $A \circ\mathbb{A}_{A}$ & $1.0$ & $56.38$ & $56.19$ \\
$A$ & $I$ & $2.0$ & $76.86$ & $55.02$ \\
$A$ & $R$ & $2.0$ & $76.03$ & $54.16$ \\
$A$ & $V$ & $2.0$ & $73.28$ & $54.38$ \\
$A$ & $A \circ\mathbb{A}_{A}$ & $2.0$ & $56.38$ & $55.89$ \\
$A$ & $I$ & $4.0$ & $76.86$ & $52.48$ \\
$A$ & $R$ & $4.0$ & $76.03$ & $50.82$ \\
$A$ & $V$ & $4.0$ & $73.28$ & $51.57$ \\
$A$ & $A \circ\mathbb{A}_{A}$ & $4.0$ & $56.38$ & $55.25$ \\
$A$ & $I$ & $8.0$ & $76.86$ & $44.76$ \\
$A$ & $R$ & $8.0$ & $76.03$ & $41.18$ \\
$A$ & $V$ & $8.0$ & $73.28$ & $43.48$ \\
$A$ & $A \circ\mathbb{A}_{A}$ & $8.0$ & $56.38$ & $52.33$ \\
$A$ & $A \circ\mathbb{A}_{A}$ & $16.0$ & $56.38$ & $41.49$ \\
$I$ & $A$ & $0.5$ & $56.48$ & $76.31$ \\
$I$ & $R$ & $0.5$ & $76.03$ & $74.63$ \\
$I$ & $V$ & $0.5$ & $73.28$ & $75.11$ \\
$I$ & $I \circ\mathbb{A}_{I}$ & $0.5$ & $76.79$ & $76.71$ \\
$I$ & $A$ & $1.0$ & $56.48$ & $75.70$ \\
$I$ & $R$ & $1.0$ & $76.03$ & $72.26$ \\
$I$ & $V$ & $1.0$ & $73.28$ & $73.29$ \\
$I$ & $I \circ\mathbb{A}_{I}$ & $1.0$ & $76.79$ & $76.45$ \\
$I$ & $A$ & $2.0$ & $56.48$ & $74.27$ \\
$I$ & $R$ & $2.0$ & $76.03$ & $68.06$ \\
$I$ & $V$ & $2.0$ & $73.28$ & $69.89$ \\
$I$ & $I \circ\mathbb{A}_{I}$ & $2.0$ & $76.79$ & $75.90$ \\
$I$ & $A$ & $4.0$ & $56.48$ & $71.20$ \\
$I$ & $R$ & $4.0$ & $76.03$ & $61.82$ \\
$I$ & $V$ & $4.0$ & $73.28$ & $64.71$ \\
$I$ & $I \circ\mathbb{A}_{I}$ & $4.0$ & $76.79$ & $74.65$ \\
$I$ & $A$ & $8.0$ & $56.48$ & $64.46$ \\
$I$ & $R$ & $8.0$ & $76.03$ & $54.70$ \\
$I$ & $V$ & $8.0$ & $73.28$ & $58.73$ \\
$I$ & $I \circ\mathbb{A}_{I}$ & $8.0$ & $76.79$ & $72.45$ \\
$I$ & $I \circ\mathbb{A}_{I}$ & $16.0$ & $76.79$ & $68.18$ \\
$R$ & $A$ & $0.5$ & $56.48$ & $75.29$ \\
$R$ & $I$ & $0.5$ & $76.86$ & $74.01$ \\
$R$ & $V$ & $0.5$ & $73.28$ & $73.22$ \\
$R$ & $R \circ\mathbb{A}_{R}$ & $0.5$ & $75.00$ & $75.71$ \\
$R$ & $R \circ\mathbb{A}_{S}$ & $0.5$ & $72.07$ & $75.03$ \\
$R$ & $A$ & $1.0$ & $56.48$ & $74.39$ \\
$R$ & $I$ & $1.0$ & $76.86$ & $71.87$ \\
$R$ & $V$ & $1.0$ & $73.28$ & $70.38$ \\
$R$ & $R \circ\mathbb{A}_{R}$ & $1.0$ & $75.00$ & $75.42$ \\
$R$ & $R \circ\mathbb{A}_{S}$ & $1.0$ & $72.07$ & $73.89$ \\
$R$ & $A$ & $2.0$ & $56.48$ & $72.36$ \\
$R$ & $I$ & $2.0$ & $76.86$ & $68.23$ \\
$R$ & $V$ & $2.0$ & $73.28$ & $65.24$ \\
$R$ & $R \circ\mathbb{A}_{R}$ & $2.0$ & $75.00$ & $74.68$ \\
$R$ & $R \circ\mathbb{A}_{S}$ & $2.0$ & $72.07$ & $71.70$ \\
$R$ & $A$ & $4.0$ & $56.48$ & $68.07$ \\
$R$ & $I$ & $4.0$ & $76.86$ & $63.06$ \\
$R$ & $V$ & $4.0$ & $73.28$ & $58.30$ \\
$R$ & $R \circ\mathbb{A}_{R}$ & $4.0$ & $75.00$ & $73.01$ \\
$R$ & $R \circ\mathbb{A}_{S}$ & $4.0$ & $72.07$ & $67.98$ \\
$R$ & $A$ & $8.0$ & $56.48$ & $58.65$ \\
$R$ & $I$ & $8.0$ & $76.86$ & $56.61$ \\
$R$ & $V$ & $8.0$ & $73.28$ & $51.20$ \\
$R$ & $R \circ\mathbb{A}_{R}$ & $8.0$ & $75.00$ & $69.59$ \\
$R$ & $R \circ\mathbb{A}_{S}$ & $8.0$ & $72.07$ & $62.50$ \\
$R$ & $R \circ\mathbb{A}_{R}$ & $16.0$ & $75.00$ & $62.54$ \\
$V$ & $A$ & $0.5$ & $56.48$ & $72.30$ \\
$V$ & $I$ & $0.5$ & $76.86$ & $70.97$ \\
$V$ & $R$ & $0.5$ & $76.03$ & $69.25$ \\
$V$ & $V \circ\mathbb{A}_{V}$ & $0.5$ & $68.14$ & $72.52$ \\
$V$ & $A$ & $1.0$ & $56.48$ & $71.19$ \\
$V$ & $I$ & $1.0$ & $76.86$ & $68.69$ \\
$V$ & $R$ & $1.0$ & $76.03$ & $65.32$ \\
$V$ & $V \circ\mathbb{A}_{V}$ & $1.0$ & $68.14$ & $71.79$ \\
$V$ & $A$ & $2.0$ & $56.48$ & $68.84$ \\
$V$ & $I$ & $2.0$ & $76.86$ & $64.78$ \\
$V$ & $R$ & $2.0$ & $76.03$ & $59.18$ \\
$V$ & $V \circ\mathbb{A}_{V}$ & $2.0$ & $68.14$ & $70.12$ \\
$V$ & $A$ & $4.0$ & $56.48$ & $63.51$ \\
$V$ & $I$ & $4.0$ & $76.86$ & $59.33$ \\
$V$ & $R$ & $4.0$ & $76.03$ & $51.14$ \\
$V$ & $V \circ\mathbb{A}_{V}$ & $4.0$ & $68.14$ & $67.02$ \\
$V$ & $A$ & $8.0$ & $56.48$ & $52.28$ \\
$V$ & $I$ & $8.0$ & $76.86$ & $51.99$ \\
$V$ & $R$ & $8.0$ & $76.03$ & $42.35$ \\
$V$ & $V \circ\mathbb{A}_{V}$ & $8.0$ & $68.14$ & $62.11$ \\
$V$ & $V \circ\mathbb{A}_{V}$ & $16.0$ & $68.14$ & $53.49$ \\
$A \circ\mathbb{A}_{A}$ & $A$ & $0.5$ & $56.48$ & $41.39$ \\
$A \circ\mathbb{A}_{A}$ & $A$ & $1.0$ & $56.48$ & $29.07$ \\
$A \circ\mathbb{A}_{A}$ & $A$ & $2.0$ & $56.48$ & $14.42$ \\
$A \circ\mathbb{A}_{A}$ & $A$ & $4.0$ & $56.48$ & $5.32$ \\
$A \circ\mathbb{A}_{A}$ & $A$ & $8.0$ & $56.48$ & $2.23$ \\
$A \circ\mathbb{A}_{A}$ & $A$ & $16.0$ & $56.48$ & $1.34$ \\
$I \circ\mathbb{A}_{I}$ & $I$ & $0.5$ & $76.86$ & $62.88$ \\
$I \circ\mathbb{A}_{I}$ & $I$ & $1.0$ & $76.86$ & $51.84$ \\
$I \circ\mathbb{A}_{I}$ & $I$ & $2.0$ & $76.86$ & $40.63$ \\
$I \circ\mathbb{A}_{I}$ & $I$ & $4.0$ & $76.86$ & $33.02$ \\
$I \circ\mathbb{A}_{I}$ & $I$ & $8.0$ & $76.86$ & $29.92$ \\
$I \circ\mathbb{A}_{I}$ & $I$ & $16.0$ & $76.86$ & $29.42$ \\
$R \circ\mathbb{A}_{R}$ & $R$ & $0.5$ & $76.03$ & $54.46$ \\
$R \circ\mathbb{A}_{R}$ & $R$ & $1.0$ & $76.03$ & $39.20$ \\
$R \circ\mathbb{A}_{R}$ & $R$ & $2.0$ & $76.03$ & $23.53$ \\
$R \circ\mathbb{A}_{R}$ & $R$ & $4.0$ & $76.03$ & $14.43$ \\
$R \circ\mathbb{A}_{R}$ & $R$ & $8.0$ & $76.03$ & $11.67$ \\
$R \circ\mathbb{A}_{R}$ & $R$ & $16.0$ & $76.03$ & $11.25$ \\
$V \circ\mathbb{A}_{V}$ & $V$ & $0.5$ & $73.28$ & $35.55$ \\
$V \circ\mathbb{A}_{V}$ & $V$ & $1.0$ & $73.28$ & $18.50$ \\
$V \circ\mathbb{A}_{V}$ & $V$ & $2.0$ & $73.28$ & $8.27$ \\
$V \circ\mathbb{A}_{V}$ & $V$ & $4.0$ & $73.28$ & $5.41$ \\
$V \circ\mathbb{A}_{V}$ & $V$ & $8.0$ & $73.28$ & $5.39$ \\
$V \circ\mathbb{A}_{V}$ & $V$ & $16.0$ & $73.28$ & $7.72$ \\
$R \circ\mathbb{A}_{S}$ & $R$ & $0.5$ & $76.03$ & $47.90$ \\
$R \circ\mathbb{A}_{S}$ & $R$ & $1.0$ & $76.03$ & $31.81$ \\
$R \circ\mathbb{A}_{S}$ & $R$ & $2.0$ & $76.03$ & $18.28$ \\
$R \circ\mathbb{A}_{S}$ & $R$ & $4.0$ & $76.03$ & $11.96$ \\
$R \circ\mathbb{A}_{S}$ & $R$ & $8.0$ & $76.03$ & $10.40$ \\
\bottomrule
\end{longtable}
\begin{longtable}[]{ccccc}
\caption{Accuracy for BIM attacks when source and target network differ.} 
\label{tab:gb-ifgs} 
\\
\toprule
Source Network & Target Network & $\epsilon$ & top-1 acc. (\%) & \dots (adversarial) \\
\midrule
\endhead
$A$ & $I$ & $0.5$ & $76.86$ & $55.04$ \\
$A$ & $R$ & $0.5$ & $76.03$ & $53.95$ \\
$A$ & $V$ & $0.5$ & $73.28$ & $54.42$ \\
$A$ & $A \circ\mathbb{A}_{A}$ & $0.5$ & $56.38$ & $54.61$ \\
$A$ & $I$ & $1.0$ & $76.86$ & $53.51$ \\
$A$ & $R$ & $1.0$ & $76.03$ & $51.69$ \\
$A$ & $V$ & $1.0$ & $73.28$ & $52.67$ \\
$A$ & $A \circ\mathbb{A}_{A}$ & $1.0$ & $56.38$ & $51.96$ \\
$A$ & $I$ & $2.0$ & $76.86$ & $48.11$ \\
$A$ & $R$ & $2.0$ & $76.03$ & $45.05$ \\
$A$ & $V$ & $2.0$ & $73.28$ & $47.17$ \\
$A$ & $A \circ\mathbb{A}_{A}$ & $2.0$ & $56.38$ & $46.29$ \\
$A$ & $I$ & $4.0$ & $76.86$ & $33.50$ \\
$A$ & $R$ & $4.0$ & $76.03$ & $29.14$ \\
$A$ & $V$ & $4.0$ & $73.28$ & $34.14$ \\
$A$ & $A \circ\mathbb{A}_{A}$ & $4.0$ & $56.38$ & $34.28$ \\
$A$ & $I$ & $8.0$ & $76.86$ & $11.08$ \\
$A$ & $R$ & $8.0$ & $76.03$ & $8.70$ \\
$A$ & $V$ & $8.0$ & $73.28$ & $12.73$ \\
$A$ & $A \circ\mathbb{A}_{A}$ & $8.0$ & $56.38$ & $14.31$ \\
$I$ & $A$ & $0.5$ & $56.48$ & $71.70$ \\
$I$ & $R$ & $0.5$ & $76.03$ & $67.32$ \\
$I$ & $V$ & $0.5$ & $73.28$ & $70.39$ \\
$I$ & $I \circ\mathbb{A}_{I}$ & $0.5$ & $76.79$ & $74.27$ \\
$I$ & $A$ & $1.0$ & $56.48$ & $67.01$ \\
$I$ & $R$ & $1.0$ & $76.03$ & $60.49$ \\
$I$ & $V$ & $1.0$ & $73.28$ & $65.72$ \\
$I$ & $I \circ\mathbb{A}_{I}$ & $1.0$ & $76.79$ & $71.67$ \\
$I$ & $A$ & $2.0$ & $56.48$ & $58.45$ \\
$I$ & $R$ & $2.0$ & $76.03$ & $49.84$ \\
$I$ & $V$ & $2.0$ & $73.28$ & $57.29$ \\
$I$ & $I \circ\mathbb{A}_{I}$ & $2.0$ & $76.79$ & $67.21$ \\
$I$ & $A$ & $4.0$ & $56.48$ & $41.90$ \\
$I$ & $R$ & $4.0$ & $76.03$ & $34.22$ \\
$I$ & $V$ & $4.0$ & $73.28$ & $44.44$ \\
$I$ & $I \circ\mathbb{A}_{I}$ & $4.0$ & $76.79$ & $58.85$ \\
$I$ & $A$ & $8.0$ & $56.48$ & $18.61$ \\
$I$ & $R$ & $8.0$ & $76.03$ & $17.36$ \\
$I$ & $V$ & $8.0$ & $73.28$ & $28.68$ \\
$I$ & $I \circ\mathbb{A}_{I}$ & $8.0$ & $76.79$ & $46.60$ \\
$R$ & $A$ & $0.5$ & $56.48$ & $68.98$ \\
$R$ & $I$ & $0.5$ & $76.86$ & $68.41$ \\
$R$ & $V$ & $0.5$ & $73.28$ & $64.93$ \\
$R$ & $R \circ\mathbb{A}_{R}$ & $0.5$ & $75.00$ & $72.31$ \\
$R$ & $R \circ\mathbb{A}_{S}$ & $0.5$ & $72.07$ & $63.43$ \\
$R$ & $A$ & $1.0$ & $56.48$ & $62.48$ \\
$R$ & $I$ & $1.0$ & $76.86$ & $62.54$ \\
$R$ & $V$ & $1.0$ & $73.28$ & $57.52$ \\
$R$ & $R \circ\mathbb{A}_{R}$ & $1.0$ & $75.00$ & $68.83$ \\
$R$ & $R \circ\mathbb{A}_{S}$ & $1.0$ & $72.07$ & $51.71$ \\
$R$ & $A$ & $2.0$ & $56.48$ & $50.59$ \\
$R$ & $I$ & $2.0$ & $76.86$ & $52.16$ \\
$R$ & $V$ & $2.0$ & $73.28$ & $46.09$ \\
$R$ & $R \circ\mathbb{A}_{R}$ & $2.0$ & $75.00$ & $63.07$ \\
$R$ & $R \circ\mathbb{A}_{S}$ & $2.0$ & $72.07$ & $35.85$ \\
$R$ & $A$ & $4.0$ & $56.48$ & $30.60$ \\
$R$ & $I$ & $4.0$ & $76.86$ & $37.46$ \\
$R$ & $V$ & $4.0$ & $73.28$ & $32.11$ \\
$R$ & $R \circ\mathbb{A}_{R}$ & $4.0$ & $75.00$ & $53.10$ \\
$R$ & $R \circ\mathbb{A}_{S}$ & $4.0$ & $72.07$ & $20.81$ \\
$R$ & $A$ & $8.0$ & $56.48$ & $10.25$ \\
$R$ & $I$ & $8.0$ & $76.86$ & $21.29$ \\
$R$ & $V$ & $8.0$ & $73.28$ & $18.94$ \\
$R$ & $R \circ\mathbb{A}_{R}$ & $8.0$ & $75.00$ & $36.36$ \\
$R$ & $R \circ\mathbb{A}_{S}$ & $8.0$ & $72.07$ & $11.40$ \\
$V$ & $A$ & $0.5$ & $56.48$ & $63.81$ \\
$V$ & $I$ & $0.5$ & $76.86$ & $64.25$ \\
$V$ & $R$ & $0.5$ & $76.03$ & $53.11$ \\
$V$ & $V \circ\mathbb{A}_{V}$ & $0.5$ & $68.14$ & $64.68$ \\
$V$ & $A$ & $1.0$ & $56.48$ & $55.92$ \\
$V$ & $I$ & $1.0$ & $76.86$ & $57.50$ \\
$V$ & $R$ & $1.0$ & $76.03$ & $41.59$ \\
$V$ & $V \circ\mathbb{A}_{V}$ & $1.0$ & $68.14$ & $56.26$ \\
$V$ & $A$ & $2.0$ & $56.48$ & $42.54$ \\
$V$ & $I$ & $2.0$ & $76.86$ & $45.69$ \\
$V$ & $R$ & $2.0$ & $76.03$ & $26.86$ \\
$V$ & $V \circ\mathbb{A}_{V}$ & $2.0$ & $68.14$ & $43.38$ \\
$V$ & $A$ & $4.0$ & $56.48$ & $23.27$ \\
$V$ & $I$ & $4.0$ & $76.86$ & $30.33$ \\
$V$ & $R$ & $4.0$ & $76.03$ & $13.02$ \\
$V$ & $V \circ\mathbb{A}_{V}$ & $4.0$ & $68.14$ & $27.12$ \\
$V$ & $A$ & $8.0$ & $56.48$ & $6.32$ \\
$V$ & $I$ & $8.0$ & $76.86$ & $15.10$ \\
$V$ & $R$ & $8.0$ & $76.03$ & $4.42$ \\
$V$ & $V \circ\mathbb{A}_{V}$ & $8.0$ & $68.14$ & $13.25$ \\
$A \circ\mathbb{A}_{A}$ & $A$ & $0.5$ & $56.48$ & $0.95$ \\
$A \circ\mathbb{A}_{A}$ & $A$ & $1.0$ & $56.48$ & $0.13$ \\
$A \circ\mathbb{A}_{A}$ & $A$ & $2.0$ & $56.48$ & $0.04$ \\
$A \circ\mathbb{A}_{A}$ & $A$ & $4.0$ & $56.48$ & $0.02$ \\
$A \circ\mathbb{A}_{A}$ & $A$ & $8.0$ & $56.48$ & $0.01$ \\
$I \circ\mathbb{A}_{I}$ & $I$ & $0.5$ & $76.86$ & $23.14$ \\
$I \circ\mathbb{A}_{I}$ & $I$ & $1.0$ & $76.86$ & $10.72$ \\
$I \circ\mathbb{A}_{I}$ & $I$ & $2.0$ & $76.86$ & $3.87$ \\
$I \circ\mathbb{A}_{I}$ & $I$ & $4.0$ & $76.86$ & $1.23$ \\
$I \circ\mathbb{A}_{I}$ & $I$ & $8.0$ & $76.86$ & $0.40$ \\
$R \circ\mathbb{A}_{R}$ & $R$ & $0.5$ & $76.03$ & $2.65$ \\
$R \circ\mathbb{A}_{R}$ & $R$ & $1.0$ & $76.03$ & $0.44$ \\
$R \circ\mathbb{A}_{R}$ & $R$ & $2.0$ & $76.03$ & $0.08$ \\
$R \circ\mathbb{A}_{R}$ & $R$ & $4.0$ & $76.03$ & $0.01$ \\
$R \circ\mathbb{A}_{R}$ & $R$ & $8.0$ & $76.03$ & $0.00$ \\
$V \circ\mathbb{A}_{V}$ & $V$ & $0.5$ & $73.28$ & $0.59$ \\
$V \circ\mathbb{A}_{V}$ & $V$ & $1.0$ & $73.28$ & $0.33$ \\
$V \circ\mathbb{A}_{V}$ & $V$ & $2.0$ & $73.28$ & $0.24$ \\
$V \circ\mathbb{A}_{V}$ & $V$ & $4.0$ & $73.28$ & $0.18$ \\
$V \circ\mathbb{A}_{V}$ & $V$ & $8.0$ & $73.28$ & $0.12$ \\
$R \circ\mathbb{A}_{S}$ & $R$ & $0.5$ & $76.03$ & $0.73$ \\
$R \circ\mathbb{A}_{S}$ & $R$ & $1.0$ & $76.03$ & $0.11$ \\
$R \circ\mathbb{A}_{S}$ & $R$ & $2.0$ & $76.03$ & $0.03$ \\
$R \circ\mathbb{A}_{S}$ & $R$ & $4.0$ & $76.03$ & $0.01$ \\
$R \circ\mathbb{A}_{S}$ & $R$ & $8.0$ & $76.03$ & $0.00$ \\
\bottomrule
\end{longtable}
\begin{longtable}[]{ccccc}
\caption{Accuracy for CW attacks when source and target network differ.} 
\label{tab:gb-cw} 
\\
\toprule
Source Network & Target Network & $\epsilon$ & top-1 acc. (\%) & \dots (adversarial) \\
\midrule
\endhead
$I$ & $I \circ\mathbb{A}_{I}$ & $0.5$ & $76.80$ & $72.90$ \\
$I$ & $I \circ\mathbb{A}_{I}$ & $1.0$ & $76.80$ & $70.08$ \\
$I$ & $I \circ\mathbb{A}_{I}$ & $2.0$ & $76.80$ & $65.54$ \\
$I$ & $I \circ\mathbb{A}_{I}$ & $4.0$ & $76.80$ & $59.46$ \\
$R$ & $R \circ\mathbb{A}_{R}$ & $0.5$ & $75.01$ & $71.15$ \\
$R$ & $R \circ\mathbb{A}_{R}$ & $1.0$ & $75.01$ & $67.98$ \\
$R$ & $R \circ\mathbb{A}_{R}$ & $2.0$ & $75.01$ & $63.42$ \\
$R$ & $R \circ\mathbb{A}_{R}$ & $4.0$ & $75.01$ & $57.05$ \\
$I \circ\mathbb{A}_{I}$ & $I$ & $0.5$ & $76.86$ & $64.36$ \\
$I \circ\mathbb{A}_{I}$ & $I$ & $1.0$ & $76.86$ & $43.53$ \\
$I \circ\mathbb{A}_{I}$ & $I$ & $2.0$ & $76.86$ & $28.18$ \\
$I \circ\mathbb{A}_{I}$ & $I$ & $4.0$ & $76.86$ & $20.69$ \\
$R \circ\mathbb{A}_{R}$ & $R$ & $0.5$ & $76.03$ & $53.83$ \\
$R \circ\mathbb{A}_{R}$ & $R$ & $1.0$ & $76.03$ & $28.23$ \\
$R \circ\mathbb{A}_{R}$ & $R$ & $2.0$ & $76.03$ & $12.76$ \\
$R \circ\mathbb{A}_{R}$ & $R$ & $4.0$ & $76.03$ & $8.50$ \\
\bottomrule
\end{longtable}
\begin{longtable}[]{ccccc}
\caption{BIM attacks on ResNet 50, mitigated with different types and strength ($\eta$) of noise.} 
\label{tab:noise-ifgs} 
\\
\toprule
Noise Type & $\epsilon$ & $\eta$ & top-1 acc. (\%) & \dots (adversarial) \\
\midrule
\endhead
Gaussian & $0.5$ & $0.5$ & $76.03$ & $2.77$ \\
Gaussian & $0.5$ & $1.0$ & $76.03$ & $3.15$ \\
Gaussian & $0.5$ & $2.0$ & $76.03$ & $4.73$ \\
Gaussian & $0.5$ & $4.0$ & $76.03$ & $11.15$ \\
Gaussian & $0.5$ & $8.0$ & $76.03$ & $29.17$ \\
Gaussian & $0.5$ & $12.0$ & $76.03$ & $41.43$ \\
Gaussian & $0.5$ & $15.0$ & $76.03$ & $46.46$ \\
Gaussian & $0.5$ & $20.0$ & $76.03$ & $50.48$ \\
Gaussian & $0.5$ & $30.0$ & $76.03$ & $48.21$ \\
Gaussian & $1.0$ & $0.5$ & $76.03$ & $0.45$ \\
Gaussian & $1.0$ & $1.0$ & $76.03$ & $0.50$ \\
Gaussian & $1.0$ & $2.0$ & $76.03$ & $0.65$ \\
Gaussian & $1.0$ & $4.0$ & $76.03$ & $1.88$ \\
Gaussian & $1.0$ & $8.0$ & $76.03$ & $11.37$ \\
Gaussian & $1.0$ & $12.0$ & $76.03$ & $24.19$ \\
Gaussian & $1.0$ & $15.0$ & $76.03$ & $31.81$ \\
Gaussian & $1.0$ & $20.0$ & $76.03$ & $40.08$ \\
Gaussian & $1.0$ & $30.0$ & $76.03$ & $43.35$ \\
Gaussian & $2.0$ & $0.5$ & $76.03$ & $0.08$ \\
Gaussian & $2.0$ & $1.0$ & $76.03$ & $0.08$ \\
Gaussian & $2.0$ & $2.0$ & $76.03$ & $0.10$ \\
Gaussian & $2.0$ & $4.0$ & $76.03$ & $0.16$ \\
Gaussian & $2.0$ & $8.0$ & $76.03$ & $1.05$ \\
Gaussian & $2.0$ & $12.0$ & $76.03$ & $5.00$ \\
Gaussian & $2.0$ & $15.0$ & $76.03$ & $10.38$ \\
Gaussian & $2.0$ & $20.0$ & $76.03$ & $20.76$ \\
Gaussian & $2.0$ & $30.0$ & $76.03$ & $32.58$ \\
Gaussian & $4.0$ & $0.5$ & $76.03$ & $0.01$ \\
Gaussian & $4.0$ & $1.0$ & $76.03$ & $0.01$ \\
Gaussian & $4.0$ & $2.0$ & $76.03$ & $0.01$ \\
Gaussian & $4.0$ & $4.0$ & $76.03$ & $0.02$ \\
Gaussian & $4.0$ & $8.0$ & $76.03$ & $0.04$ \\
Gaussian & $4.0$ & $12.0$ & $76.03$ & $0.10$ \\
Gaussian & $4.0$ & $15.0$ & $76.03$ & $0.30$ \\
Gaussian & $4.0$ & $20.0$ & $76.03$ & $1.85$ \\
Gaussian & $4.0$ & $30.0$ & $76.03$ & $11.53$ \\
Gaussian & $8.0$ & $0.5$ & $76.03$ & $0.00$ \\
Gaussian & $8.0$ & $1.0$ & $76.03$ & $0.00$ \\
Gaussian & $8.0$ & $2.0$ & $76.03$ & $0.00$ \\
Gaussian & $8.0$ & $4.0$ & $76.03$ & $0.00$ \\
Gaussian & $8.0$ & $8.0$ & $76.03$ & $0.00$ \\
Gaussian & $8.0$ & $12.0$ & $76.03$ & $0.00$ \\
Gaussian & $8.0$ & $15.0$ & $76.03$ & $0.01$ \\
Gaussian & $8.0$ & $20.0$ & $76.03$ & $0.01$ \\
Gaussian & $8.0$ & $30.0$ & $76.03$ & $0.26$ \\
Sign & $0.5$ & $0.5$ & $76.03$ & $2.77$ \\
Sign & $0.5$ & $1.0$ & $76.03$ & $3.14$ \\
Sign & $0.5$ & $2.0$ & $76.03$ & $4.81$ \\
Sign & $0.5$ & $4.0$ & $76.03$ & $11.25$ \\
Sign & $0.5$ & $8.0$ & $76.03$ & $29.59$ \\
Sign & $0.5$ & $12.0$ & $76.03$ & $41.88$ \\
Sign & $0.5$ & $15.0$ & $76.03$ & $47.01$ \\
Sign & $0.5$ & $20.0$ & $76.03$ & $50.73$ \\
Sign & $0.5$ & $30.0$ & $76.03$ & $47.74$ \\
Sign & $1.0$ & $0.5$ & $76.03$ & $0.45$ \\
Sign & $1.0$ & $1.0$ & $76.03$ & $0.47$ \\
Sign & $1.0$ & $2.0$ & $76.03$ & $0.65$ \\
Sign & $1.0$ & $4.0$ & $76.03$ & $1.94$ \\
Sign & $1.0$ & $8.0$ & $76.03$ & $11.61$ \\
Sign & $1.0$ & $12.0$ & $76.03$ & $24.76$ \\
Sign & $1.0$ & $15.0$ & $76.03$ & $32.74$ \\
Sign & $1.0$ & $20.0$ & $76.03$ & $40.71$ \\
Sign & $1.0$ & $30.0$ & $76.03$ & $43.30$ \\
Sign & $2.0$ & $0.5$ & $76.03$ & $0.08$ \\
Sign & $2.0$ & $1.0$ & $76.03$ & $0.08$ \\
Sign & $2.0$ & $2.0$ & $76.03$ & $0.09$ \\
Sign & $2.0$ & $4.0$ & $76.03$ & $0.16$ \\
Sign & $2.0$ & $8.0$ & $76.03$ & $1.13$ \\
Sign & $2.0$ & $12.0$ & $76.03$ & $5.38$ \\
Sign & $2.0$ & $15.0$ & $76.03$ & $11.19$ \\
Sign & $2.0$ & $20.0$ & $76.03$ & $21.74$ \\
Sign & $2.0$ & $30.0$ & $76.03$ & $33.49$ \\
Sign & $4.0$ & $0.5$ & $76.03$ & $0.01$ \\
Sign & $4.0$ & $1.0$ & $76.03$ & $0.01$ \\
Sign & $4.0$ & $2.0$ & $76.03$ & $0.01$ \\
Sign & $4.0$ & $4.0$ & $76.03$ & $0.02$ \\
Sign & $4.0$ & $8.0$ & $76.03$ & $0.04$ \\
Sign & $4.0$ & $12.0$ & $76.03$ & $0.12$ \\
Sign & $4.0$ & $15.0$ & $76.03$ & $0.38$ \\
Sign & $4.0$ & $20.0$ & $76.03$ & $2.25$ \\
Sign & $4.0$ & $30.0$ & $76.03$ & $13.05$ \\
Sign & $8.0$ & $0.5$ & $76.03$ & $0.00$ \\
Sign & $8.0$ & $1.0$ & $76.03$ & $0.00$ \\
Sign & $8.0$ & $2.0$ & $76.03$ & $0.00$ \\
Sign & $8.0$ & $4.0$ & $76.03$ & $0.00$ \\
Sign & $8.0$ & $8.0$ & $76.03$ & $0.00$ \\
Sign & $8.0$ & $12.0$ & $76.03$ & $0.00$ \\
Sign & $8.0$ & $15.0$ & $76.03$ & $0.01$ \\
Sign & $8.0$ & $20.0$ & $76.03$ & $0.01$ \\
Sign & $8.0$ & $30.0$ & $76.03$ & $0.44$ \\
Uniform & $0.5$ & $0.5$ & $76.03$ & $2.72$ \\
Uniform & $0.5$ & $1.0$ & $76.03$ & $2.81$ \\
Uniform & $0.5$ & $2.0$ & $76.03$ & $3.37$ \\
Uniform & $0.5$ & $4.0$ & $76.03$ & $5.57$ \\
Uniform & $0.5$ & $8.0$ & $76.03$ & $13.82$ \\
Uniform & $0.5$ & $12.0$ & $76.03$ & $24.69$ \\
Uniform & $0.5$ & $15.0$ & $76.03$ & $31.86$ \\
Uniform & $0.5$ & $20.0$ & $76.03$ & $40.42$ \\
Uniform & $0.5$ & $30.0$ & $76.03$ & $49.01$ \\
Uniform & $1.0$ & $0.5$ & $76.03$ & $0.44$ \\
Uniform & $1.0$ & $1.0$ & $76.03$ & $0.44$ \\
Uniform & $1.0$ & $2.0$ & $76.03$ & $0.49$ \\
Uniform & $1.0$ & $4.0$ & $76.03$ & $0.76$ \\
Uniform & $1.0$ & $8.0$ & $76.03$ & $2.69$ \\
Uniform & $1.0$ & $12.0$ & $76.03$ & $8.04$ \\
Uniform & $1.0$ & $15.0$ & $76.03$ & $13.55$ \\
Uniform & $1.0$ & $20.0$ & $76.03$ & $23.04$ \\
Uniform & $1.0$ & $30.0$ & $76.03$ & $37.01$ \\
Uniform & $2.0$ & $0.5$ & $76.03$ & $0.08$ \\
Uniform & $2.0$ & $1.0$ & $76.03$ & $0.08$ \\
Uniform & $2.0$ & $2.0$ & $76.03$ & $0.08$ \\
Uniform & $2.0$ & $4.0$ & $76.03$ & $0.10$ \\
Uniform & $2.0$ & $8.0$ & $76.03$ & $0.20$ \\
Uniform & $2.0$ & $12.0$ & $76.03$ & $0.63$ \\
Uniform & $2.0$ & $15.0$ & $76.03$ & $1.43$ \\
Uniform & $2.0$ & $20.0$ & $76.03$ & $4.51$ \\
Uniform & $2.0$ & $30.0$ & $76.03$ & $15.73$ \\
Uniform & $4.0$ & $0.5$ & $76.03$ & $0.01$ \\
Uniform & $4.0$ & $1.0$ & $76.03$ & $0.01$ \\
Uniform & $4.0$ & $2.0$ & $76.03$ & $0.01$ \\
Uniform & $4.0$ & $4.0$ & $76.03$ & $0.01$ \\
Uniform & $4.0$ & $8.0$ & $76.03$ & $0.02$ \\
Uniform & $4.0$ & $12.0$ & $76.03$ & $0.03$ \\
Uniform & $4.0$ & $15.0$ & $76.03$ & $0.04$ \\
Uniform & $4.0$ & $20.0$ & $76.03$ & $0.09$ \\
Uniform & $4.0$ & $30.0$ & $76.03$ & $0.83$ \\
Uniform & $8.0$ & $0.5$ & $76.03$ & $0.00$ \\
Uniform & $8.0$ & $1.0$ & $76.03$ & $0.00$ \\
Uniform & $8.0$ & $2.0$ & $76.03$ & $0.00$ \\
Uniform & $8.0$ & $4.0$ & $76.03$ & $0.00$ \\
Uniform & $8.0$ & $8.0$ & $76.03$ & $0.00$ \\
Uniform & $8.0$ & $12.0$ & $76.03$ & $0.00$ \\
Uniform & $8.0$ & $15.0$ & $76.03$ & $0.00$ \\
Uniform & $8.0$ & $20.0$ & $76.03$ & $0.00$ \\
Uniform & $8.0$ & $30.0$ & $76.03$ & $0.01$ \\
\bottomrule
\end{longtable}

\end{document}